\documentclass[10pt,journal,compsoc]{IEEEtran}

\ifCLASSOPTIONcompsoc
  \usepackage[nocompress]{cite}
\else
  \usepackage{cite}
\fi

\ifCLASSINFOpdf

\else

\fi
\usepackage{caption}
\usepackage{times,amsmath,epsfig}

\usepackage{algorithm}

\usepackage{algorithmic}
\usepackage{booktabs}
\usepackage{amsthm}
\usepackage{graphicx}
\usepackage{epstopdf}
\usepackage{amssymb}
\newtheorem*{definition}{Definition}

\newcommand{\paratitle}[1]{\vspace{0.8ex}\noindent \textbf{#1}}
\newcommand{\tabincell}[2]{\begin{tabular}{@{}#1@{}}#2\end{tabular}}
\usepackage{multirow}
\usepackage{array}
\usepackage{subfigure}
\usepackage{color}
\usepackage{colortbl}
\usepackage[framemethod=tikz]{mdframed}
\usepackage{hyperref}
\hypersetup{
    linkcolor=black,
    filecolor=black,
    urlcolor=blue,
    citecolor=black,
}

\begin{document}

\title{Entity Linking Meets Deep Learning:\\ Techniques and Solutions}

\author{
Wei Shen, Yuhan Li, Yinan Liu, Jiawei Han, \IEEEmembership{Fellow,~IEEE}, Jianyong Wang, \IEEEmembership{Fellow,~IEEE}, Xiaojie Yuan
\IEEEcompsocitemizethanks{
\IEEEcompsocthanksitem W. Shen, Y. Li, Y. Liu and X. Yuan are with the TKLNDST, College of Computer Science, Nankai University, Tianjin 300350, China. E-mail: \{shenwei, yuanxj\}@nankai.edu.cn,
\{yuhanli, liuyn\}@mail.nankai.edu.cn.
\IEEEcompsocthanksitem J. Han is with the Department of Computer Science, University of Illinois at Urbana-Champaign,
Urbana, IL 61801, USA. E-mail: hanj@illinois.edu.
\IEEEcompsocthanksitem J. Wang is with the Department of Computer Science and Technology, Tsinghua University, Beijing 100084, China, and also with the Jiangsu Collaborative Innovation Center for Language Ability, Jiangsu Normal University, Xuzhou 221009, China. E-mail: jianyong@tsinghua.edu.cn.
}
\thanks{To appear in IEEE TKDE.}
}

\markboth{IEEE TRANSACTIONS ON KNOWLEDGE AND DATA ENGINEERING, 2021}
{Shen \etal: Entity Linking Meets Deep Learning: Techniques and Solutions}

\IEEEtitleabstractindextext{%
\begin{abstract}

Entity linking (EL) is the process of linking entity mentions appearing in web text with their corresponding entities in a knowledge base. EL plays an important role in the fields of knowledge engineering and data mining, underlying a variety of downstream applications such as knowledge base population, content analysis, relation extraction, and question answering. In recent years, deep learning (DL), which has achieved tremendous success in various domains, has also been leveraged in EL methods to surpass traditional machine learning based methods and yield the state-of-the-art performance. In this survey, we present a comprehensive review and analysis of existing DL based EL methods. First of all, we propose a new taxonomy, which organizes existing DL based EL methods using three axes: embedding, feature, and algorithm. Then we systematically survey the representative EL methods along the three axes of the taxonomy. Later, we introduce ten commonly used EL data sets and give a quantitative performance analysis of DL based EL methods over these data sets. Finally, we discuss the remaining limitations of existing methods and highlight some promising future directions.

\end{abstract}

\begin{IEEEkeywords}
Entity linking, deep learning, entity disambiguation, knowledge base
\end{IEEEkeywords}}

\maketitle

\IEEEdisplaynontitleabstractindextext

\IEEEpeerreviewmaketitle

\IEEEraisesectionheading{\section{Introduction}\label{section1}}

\IEEEPARstart{T}{he} rapid growth of web data contains an overwhelming amount of information and knowledge. Natural language is one of the most important forms of web data. However, natural language is ambiguous, especially for named entities which appear in it frequently. In addition, with the development of information extraction (IE) techniques, a growing number of high-quality, large-scale, and machine-readable knowledge bases (KBs) have been constructed recently, such as YAGO \cite{suchanek2007yago}, DBpedia \cite{auer2007dbpedia}, Freebase \cite{bollacker2008freebase}, and Probase \cite{wu2012probase}. These KBs contain tens of millions of named entities and billions of relational facts between named entities. Bridging web text data and KBs is very helpful for understanding the ambiguous natural language on the web and enriching the existing KBs. To achieve this goal, entity linking (EL) is a fundamental task which needs to be solved.

\begin{figure}[t]
  \centering
  \includegraphics[width=0.5\textwidth]{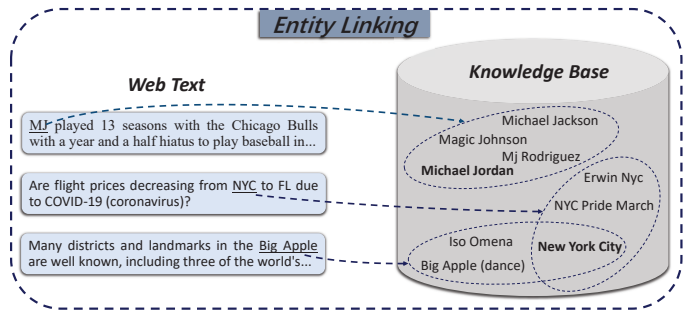}
  \caption{An illustration for the task of entity linking. Entity mentions detected from web text are \underline{underlined}; Candidate mapping entities in a knowledge base for each entity mention are shown via a dashed arrow line and circle; Their correct mapping entities (a.k.a. gold mapping entities) are in \textbf{boldface}.}
  \label{EL_illustration}
  \vspace{-0.3cm}
\end{figure}

\paratitle{What is entity linking?}
Entity linking is the task to link entity mentions appearing in web text with their corresponding entities in a KB \cite{shen2014entity}. Figure \ref{EL_illustration} shows an illustration for the entity linking task. It acts an important pre-processing step for many downstream applications, such as question answering \cite{zhang2016joint}, relation extraction \cite{lin2016neural}, knowledge base population (KBP) \cite{ji2011knowledge}, and content analysis \cite{michelson2010discovering}. For example, KBP is the task of enriching existing KBs with newly extracted facts from the text. Before the enrichment, an EL system is needed to map entity mentions associated with facts to their corresponding named entities in a KB.

Entity linking is challenging due to the name variation and entity ambiguity. On one hand, a named entity may have many different surface forms (e.g., full name, partial name, nickname, alias, and abbreviation). As the example shown in Figure \ref{EL_illustration}, entity mentions ``NYC'' and ``Big Apple'' are the abbreviation and nickname of the named entity ``New York City'', respectively. On the other hand, an entity mention may refer to many different named entities. For the example in Figure \ref{EL_illustration}, the entity mention ``MJ'' may refer to an American professional basketball player named ``Michael Jordan'', an American recording artist named ``Michael Jackson'', an American singer named ``Mj Rodriguez'', or many other named entities which could be referred to as ``MJ''.

\paratitle{Traditional machine learning based entity linking.} Many traditional machine learning (ML) based EL methods have been comprehensively reviewed in a survey \cite{shen2014entity}. These ML based EL methods mainly leverage manually designed features of entity popularity, local context compatibility, and document-level global coherence of referring entities. Their entity ranking techniques can be broadly divided into two categories: unsupervised and supervised ranking methods. Unsupervised ranking methods include vector space model based methods \cite{cucerzan2007large} and information retrieval based methods \cite{varma2009iiit}. Supervised ranking methods mainly include binary classification methods \cite{chen2011collaborative}, learning to rank methods \cite{ratinov2011local}, probabilistic methods \cite{shen2017shine+}, and graph-based methods \cite{hoffart2011robust}.

Most traditional ML based EL methods follow the popular two-step procedure, which firstly extracts some hand-crafted features and then feeds those features to an entity ranking method to make the final linking predictions. However, these methods have two limitations: (1) features that lead to good performance require a lot of careful and tedious feature engineering; (2) generalizing the trained entity linking model to other KBs or domains is difficult due to the strong dependence on the specific KB and domain knowledge in the process of designing features. 

\paratitle{Deep learning based entity linking.}
With the success of deep learning in various domains, deep learning (DL) based EL methods have been proposed and attracted significant attention recently \cite{sevgili2020neural}. Compared with traditional machine learning, deep learning can automatically learn important features and is more transferable from one domain to another. For example, DL based EL methods are able to obtain vector representations of the words via a pre-trained language model such as Word2Vec \cite{mikolov2013efficient,mikolov2013distributed} and GloVe \cite{pennington2014glove}. This neural encoding can be directly used as semantic and syntactic features, which are then fed into neural network architectures, such as recurrent neural networks (RNNs) \cite{hochreiter1997long,cho2014learning}, convolutional neural networks (CNNs) \cite{lecun1998gradient}, attention \cite{bahdanau2015neural}, and Transformers \cite{vaswani2017attention} to capture more sophisticated long-distance feature representations.

\begin{figure}[t]
  \centering
  \includegraphics[width=0.43\textwidth]{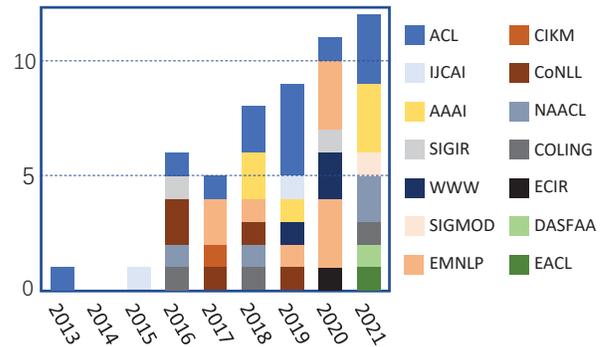}
  \vspace{-0.1cm}
  \caption{Statistics of DL based EL publications in peer-reviewed venues.}
  \label{fig:conference}
  \vspace{-0.4cm}
\end{figure}

\paratitle{Motivations.}
In the past few years, DL based EL methods with minimal feature engineering have been flourishing as shown in Figure \ref{fig:conference}. Deep learning has become the mainstream framework for the entity linking methods that achieve the state-of-the-art performance. In the previous comprehensive survey \cite{shen2014entity} of entity linking published in 2015, we reviewed and analyzed a variety of traditional ML based EL systems, which are mainly based on manually designed features without the use of deep learning. This prompts us to write this survey to systematically review and summarize the current status of deep learning techniques for entity linking. Our intended audience includes researchers, developers, and practitioners in related fields who are interested in entity linking. We hope that this survey paper will help them get a clear overall picture of current DL based EL methods and guide them to the right way to start with entity linking research and practice.

\paratitle{Contributions.} Overall, the contributions of this survey can be summarized as follows.

\begin{itemize}
  \item \textit{\textbf{A comprehensive review.}} We conduct a comprehensive survey to present a thorough overview and analysis of DL based EL methods. 
  \item \textit{\textbf{A new taxonomy.}} We propose a new taxonomy, which organizes the techniques for ranking candidate entities in existing DL based EL methods using three axes: (1) embedding; (2) feature; (3) algorithm.
  \item \textit{\textbf{A quantitative analysis.}} We present a quantitative analysis on the performance of the DL based EL methods on a variety of data sets.
  \item \textit{\textbf{Some future directions.}} We discuss the remaining limitations of existing entity linking methods and point out possible future directions.
\end{itemize}
 
The remainder of this survey is organized as follows. We first give a formal formulation of the entity linking task in Section \ref{problemoverview}. Then we explain our new taxonomy of DL based EL methods in Section \ref{taxonomy}. We present a comprehensive survey along the three axes of the new taxonomy (i.e., embedding, feature, and algorithm) in Sections \ref{embeddings}, \ref{features}, and \ref{algorithm} respectively. Section \ref{evaluation} introduces ten commonly used entity linking data sets as well as evaluation metrics, and presents a quantitative performance analysis of the DL based EL methods. Section \ref{future} discusses the main limitations and future directions, and in Section \ref{conclusion} we finally give a conclusion.

\vspace{-2.5mm}
\section{Problem Overview}
\label{problemoverview}
\vspace{-1mm}

An entity mention is a token sequence in text which potentially refers to some named entity. A named entity is a word or a phrase that is explicitly defined by KBs with a unique identifier, and it could be a real-world object (e.g., organization and individual) or an abstract concept (e.g., song and event). Most of the existing EL works assume that entity mention boundaries are provided by users or mention detection systems, such as Named Entity Recognition (NER) tools, which can identify the boundaries of named entities in text automatically. Then we formally state the task of EL as follows.

\begin{definition}[\textbf{Entity Linking}]
   Given a document $D$ containing a set of recognized entity mentions $M=\{m_1, m_2, ..., m_{|M|}\}$ and a target KB containing a set of named entities $E = \{e_1, e_2, ..., e_{|E|}\}$, the goal is to map each entity mention $m_i$ in $M$ to its gold mapping entity $e_i$ in $E$. 
\end{definition}
 
It is possible that some entity mention's gold mapping entity does not exist in the target KB, which is called ``unlinkable entity mention''. A special label NIL is usually used to describe this kind of unlinkable entity mention. 

As surveyed in \cite{shen2014entity}, a typical EL system often consists of three stages: (1) \textit{candidate entity generation} stage can generate a candidate mapping entity set for each entity mention using a name dictionary and web information; (2) \textit{candidate entity ranking} stage can leverage different kinds of evidence to rank the candidate entities; (3) \textit{unlinkable mention prediction} stage can validate whether the entity mention should be labeled as NIL. As the techniques used in the first and third stages have not changed much in recent years, in this survey we mainly concentrate on the \textit{candidate entity ranking} stage and discuss how different neural architectures and features help to rank entities. For the technical details of approaches used in the first and third stages, you could refer to our previous survey \cite{shen2014entity}.

\vspace{-2.5mm}
\section{A Taxonomy of DL Based EL Methods}
\label{taxonomy}
\vspace{-1mm}

In this section, we first introduce basic concepts of DL based EL methods and explain why EL methods can benefit from DL. Next, we give a brief introduction of the newly proposed taxonomy.

\vspace{-2.5mm}
\subsection{Why Choose Deep Learning?}
\label{whydl}
DL is an artificial intelligence function that imitates the working mechanism of the human brain in processing information and creating patterns. It utilizes a cascade of multiple layers of non-linear processing units (i.e., neurons) for feature extraction and transformation by adjusting the connection weights between units, which can be regarded as resembling the learning behavior of a human brain. DL models have been applied widely and successfully to numerous tasks in the fields of computer vision (CV) and natural language process (NLP) since the DL based model AlexNet \cite{krizhevsky2012imagenet} won the ImageNet competition by a big margin in 2012.

The most basic model in DL corresponds to a fully connected layer \cite{mudgal2018deep}. It takes vectors $(\mathbf{x}_1, \mathbf{x}_2, ..., \mathbf{x}_n)$ from the input layer as input, and outputs a value $y = f(\sum_{i=1}^{n}\mathcal{W}_i\mathbf{x}_i + b)$ by a non-linear activation function $f(\cdot)$ at the output layer, where $\mathcal{W}_i$ are weights, $b$ is the bias, and the common choice of $f(\cdot)$ is \textit{tanh} function, \textit{sigmoid} function, or \textit{ReLU} function \cite{glorot2011deep}. Multi-layer neural network is simply a generalization of this basic model where neural network layers are stacked in sequence. A multi-layer neural network is also called a multi-layer perceptron (MLP).

Compared with traditional ML, DL has three main strengths to solve the EL task. First and foremost, DL can utilize the given documents and KBs to automatically discover multiple levels of distributed representations to help disambiguation without human intervention \cite{bengio2013deep}, while traditional ML based methods require considerable feature engineering and analysis. Second, DL is more transferable, which means that deep neural networks can learn more transferable representations that disentangle the exploratory factors of variations underlying the data samples and group features hierarchically in accordance with their relatedness to invariant factors \cite{wang2018deep}. Finally, DL is able to learn feature representations and perform classification or regression in an end-to-end style, which possibly motivates complex and advanced EL methods. Along with these advantages of DL, a lot of DL based EL methods have been proposed in recent years and achieved the state-of-the-art performance. 

\vspace{-2.5mm}
\subsection{Basic Structure of Our Taxonomy}
Previous survey \cite{shen2014entity} roughly divides the candidate entity ranking methods into two categories: supervised ranking methods and unsupervised ranking methods. Different from the previous survey, we propose a new taxonomy of the techniques in the \textit{candidate entity ranking} stage of DL based EL methods, which contains three steps as follows:

\begin{itemize}

  \item \textit{\textbf{Embedding.}} Embedding can use low-dimensional and dense vectors to implicitly represent the semantic and syntactic properties of natural language, which is also called distributed representation. Generally, DL models need embeddings as input \cite{collobert2011natural}. Thus, a key point in applying DL to EL is the embeddings of different inputs (e.g., mention surface form, entity context, and entity description). Lots of embedding techniques have been utilized by existing DL based EL methods, and we divide them into four categories: (1) word embedding; (2) mention embedding; (3) entity embedding; (4) alignment embedding.
  
  \item \textit{\textbf{Feature.}} Features intend to measure the similarity between the entity mention and the candidate entity in different aspects in the form of a vector or a score. According to the previous survey \cite{shen2014entity}, almost all traditional EL methods are based on quantitative features of mention-entity popularity, mention-entity similarity, and entity-entity topical coherence. These features are still widely utilized by researchers until now, and they can benefit from DL according to the following two aspects: (1) features can be directly learned by neural models without manual feature engineering. For example, Wu et al. \cite{wu2020dynamic} learned an entity-entity topical coherence feature vector for each candidate entity automatically via a dynamic Graph Convolutional Network architecture; (2) embeddings introduced above can be leveraged to generate diverse features. For example, unlike traditional methods preferring to represent the context as a bag-of-words, many DL methods could leverage the embedding of text to calculate the similarity between mention context and entity description as a context similarity feature. In this survey, we concentrate on the following five principal features for entity linking: (1) prior popularity; (2) surface form similarity; (3) type similarity; (4) context similarity; (5) topical coherence. 
  
  \item \textit{\textbf{Algorithm.}} Algorithm is the final step in the \textit{candidate entity ranking} stage. It takes features as input and outputs the final entity linking result. The commonly used algorithms in DL based EL methods include MLP, PageRank, graph regularization, and reinforcement learning. The output of the algorithm is a score list of all candidate entities for each entity mention. We summarize such algorithms into three groups, namely, MLP, graph-based algorithms, and reinforcement learning (RL).

\end{itemize}

Although the implementation details of DL based EL methods may vary considerably, their general steps for ranking entities are very similar. Given an entity mention in a document and a set of candidate entities, various embeddings are generated at first. Then features can be calculated using embeddings. Finally, features are fed into algorithms to rank the candidate entities and get the final linking result. It is noted that these three steps are in a sequence manner, which means the output of the previous step is used as the input of the current step. In the following, we review the main techniques used in these three steps in detail. Table \ref{tab:tabsurvey} summarizes the step choices of DL based EL methods according to the proposed taxonomy.

\begin{table*}[t]
  \centering
  \caption{Summary of the step choices of DL based EL methods according to our proposed taxonomy. Due to the limited space, ``PP'' refers to the prior popularity feature, ``SFS'' refers to the surface form similarity feature, ``TS'' refers to the type similarity feature, ``CS'' refers to the context similarity feature, and ``TC'' refers to the topical coherence feature. ``-" in the column of Algorithms means the corresponding model uses a relatively simple algorithm to select the mapping entity, such as a linear combination of features.} 
  \label{tab:tabsurvey}
    \scalebox{0.88}{
  \begin{tabular}{c|c|c|c|c|c|c|c|c|c|c}
    \hline
    \multirow{2}{*}{\textbf{Model}} & \multicolumn{4}{c|}{\textbf{Embeddings}} & \multicolumn{5}{c|}{\textbf{Features}} & \multirow{2}{*}{\textbf{Algorithms}} \\ \cline{2-10}
    & \textbf{Word}      & \textbf{Mention}      & \textbf{Entity}      &    \textbf{Alignment}       & \textbf{PP} &  \textbf{SFS}  & \textbf{TS} & \textbf{CS} & \textbf{TC} &  \\  \hline
    \textbf{He et al.} (ACL 2013) \cite{he2013learning} &  &  & description &  &  &  &  &  \checkmark &  & - \\ \hline
    \textbf{Sun et al.} (IJCAI 2015) \cite{sun2015modeling} & learned & \checkmark & surface form, type &  &  &  &  & \checkmark &  & - \\ \hline
    \textbf{DSRM} (arXiv 2015) \cite{huang2015leveraging} &  &  & description, context, type &  & \checkmark & \checkmark &  &  & \checkmark & graph-based \\ \hline
    \textbf{Globerson et al.} (ACL 2016) \cite{globerson2016collective} &  &  &  &  & \checkmark & \checkmark &  &  & \checkmark & - \\ \hline
    \textbf{Zwicklbauer et al.} (SIGIR 2016) \cite{zwicklbauer2016robust} & pre-trained &  & description, context &  & \checkmark &  &  & \checkmark & \checkmark & graph-based \\ \hline
    \textbf{EDKate} (CoNLL 2016) \cite{fang2016entity} & learned &  & context & \checkmark & \checkmark & \checkmark &  & \checkmark & \checkmark & - \\ \hline
    \textbf{Yamada et al.} (CoNLL 2016) \cite{yamada2016joint} & learned &  & context & \checkmark & \checkmark & \checkmark &  &  \checkmark & \checkmark & - \\ \hline
    \textbf{Francis-Landau et al.} (NAACL 2016) \cite{francis2016capturing} & learned & \checkmark & surface form, description &  & \checkmark & \checkmark & & \checkmark &  & - \\ \hline
    \textbf{Nguyen et al.} (COLING 2016) \cite{nguyen2016joint} & learned & \checkmark & surface form, description &  & \checkmark & \checkmark &  & \checkmark & \checkmark & -\\ \hline
    \textbf{Cao et al.} (ACL 2017) \cite{cao2017bridge} & learned & \checkmark & context & \checkmark & \checkmark &  &  & \checkmark & \checkmark & - \\ \hline
    \textbf{Gupta et al.} (EMNLP 2017) \cite{gupta2017entity} & pre-trained &  & description, type &  & \checkmark &  &  & \checkmark &  & -\\ \hline
    \textbf{Deep-ED} (EMNLP 2017) \cite{ganea2017deep} & pre-trained &  & description, context &  & \checkmark &  &  & \checkmark & \checkmark & MLP \\ \hline
    \textbf{NeuPL} (CIKM 2017) \cite{phan2017neupl} & learned &  & description, context & \checkmark & \checkmark & \checkmark &  & \checkmark & \checkmark & - \\ \hline
    \textbf{Eshel et al.} (CoNLL 2017) \cite{eshel2017named} & learned &  & context & \checkmark & \checkmark & \checkmark &  & \checkmark &  & MLP \\ \hline
    \textbf{MR-Deep-ED} (ACL 2018) \cite{le2018improving} & pre-trained &  & description, context &  & \checkmark &  &  & \checkmark & \checkmark & MLP \\ \hline 
    \textbf{Moon et al.} (ACL 2018) \cite{moon2018multimodal} & pre-trained &  & context &  &  & \checkmark &  & \checkmark &  & - \\ \hline
    \textbf{Sil et al.} (AAAI 2018) \cite{sil2017neural} & learned &  & description &  & \checkmark &  &  & \checkmark &  & MLP \\ \hline
    \textbf{DeepType} (AAAI 2018) \cite{raiman2018deeptype} &  &  &  &  & \checkmark &  & \checkmark &  &  & - \\ \hline
    \textbf{Mueller and Durrett} (EMNLP 2018) \cite{mueller2018effective} & learned &  & context & \checkmark & \checkmark & \checkmark &  & \checkmark &  & MLP \\ \hline
    \textbf{Kolitsas et al.} (CoNLL 2018) \cite{kolitsas2018end} & pre-trained & \checkmark & description, context &  & \checkmark &  &  & \checkmark & \checkmark & MLP \\ \hline
    \textbf{SGTB-BiBSG} (NAACL 2018) \cite{yang2018collective} & pre-trained &  & description, context &  & \checkmark &  &  & \checkmark & \checkmark & -\\ \hline
    \textbf{NCEL} (COLING 2018) \cite{cao2018neural} & learned & \checkmark & context & \checkmark & \checkmark & \checkmark &  & \checkmark & \checkmark & MLP \\ \hline
    \textbf{Le and Titov} (ACL 2019) \cite{le2019distant} & pre-trained &  & type &  &  &  &  & \checkmark &  & MLP \\ \hline
    \textbf{Le and Titov} (ACL 2019) \cite{le2019boosting} & pre-trained &  & description, context &  & \checkmark &  &  & \checkmark & \checkmark & MLP \\ \hline
    \textbf{Logeswaran et al.} (ACL 2019) \cite{logeswaran2019zero} &  &  &  &  &  &  &  & \checkmark & &- \\ \hline
    \textbf{Sevgili et al.} (ACL 2019) \cite{sevgili2019improving} &  &  & description, context &  &  &  &  & \checkmark &  & MLP \\ \hline
    \textbf{RRWEL} (IJCAI 2019) \cite{xue2019neural} & learned & \checkmark & surface form, description &  & \checkmark & \checkmark &  & \checkmark & \checkmark & graph-based \\ \hline
    \textbf{RLEL} (WWW 2019) \cite{fang2019joint} & pre-trained &  & description, context &  & &  &  & \checkmark & \checkmark & RL \\ \hline
    \textbf{DCA} (EMNLP 2019) \cite{yang2019learning} & pre-trained & \checkmark & surface form, description, context &  & \checkmark & \checkmark & \checkmark & \checkmark & \checkmark & MLP, RL \\ \hline
    \textbf{Gillick et al.} (CoNLL 2019) \cite{gillick2019learning} & pre-trained &  & description &  & \checkmark &  &  & \checkmark &  & MLP \\ \hline
    \textbf{E-ELMo} (arXiv 2019) \cite{shahbazi2019entity} & learned &  & context &  & \checkmark & \checkmark &  & \checkmark &  & MLP \\ \hline
    \textbf{FGS2EE} (ACL 2020) \cite{hou2020improving} & pre-trained &  & description, context, type &  & \checkmark &  &  & \checkmark & \checkmark & MLP \\ \hline
    \textbf{ET4EL} (AAAI 2020) \cite{onoe2020fine} & learned & \checkmark &  &  & \checkmark &  & \checkmark &  &  & - \\ \hline
    \textbf{Chen et al.} (AAAI 2020) \cite{chen2020improving} & pre-trained &  & description, context, type &  & \checkmark &  &  & \checkmark & \checkmark & MLP \\ \hline
    \textbf{REL} (SIGIR 2020) \cite{van2020rel} & learned &  & context & \checkmark & \checkmark &  &  & \checkmark & \checkmark & MLP \\ \hline
    \textbf{SeqGAT} (WWW 2020) \cite{fang2020high} &  &  & description &  & \checkmark & \checkmark &  & \checkmark & \checkmark & MLP \\ \hline
    \textbf{DGCN} (WWW 2020) \cite{wu2020dynamic} &  & \checkmark & description, context, type &  & \checkmark &  &  & \checkmark & \checkmark & MLP \\ \hline
    \textbf{BLINK} (EMNLP 2020) \cite{wu2020scalable} &  &  & description &  &  &  &  & \checkmark &  &- \\ \hline
    \textbf{ELQ} (EMNLP 2020) \cite{li2020efficient} &  & \checkmark & description &  &  &  &  & \checkmark &  &- \\ \hline
    \textbf{GNED} (KBS 2020) \cite{hu2020graph} & pre-trained &  & description, context &  & \checkmark &  &  & \checkmark & \checkmark & MLP \\ \hline
    \textbf{JMEL} (ECIR 2020) \cite{adjali2020ecir} & learned &  &  &  &  &  &  & \checkmark & & MLP \\ \hline
    \textbf{Yamada et al.} (arXiv 2020) \cite{yamada2020global} &  & \checkmark & context &  &  &  &  & \checkmark & \checkmark &- \\ \hline
    \textbf{M3} (AAAI 2021) \cite{gu2021read} &  &  &  &  &  &  &  & \checkmark & \checkmark & -\\ \hline
    \textbf{Bi-MPR} (AAAI 2021) \cite{tang2021bidirectional} &  &  & description &  &  &  &  & \checkmark & & MLP \\ \hline
    \textbf{Chen et al.} (AAAI 2021) \cite{chen2021lightweight} & learned & \checkmark & surface form &  & \checkmark & \checkmark &  & \checkmark & \checkmark & MLP \\ \hline
    \textbf{CHOLAN} (EACL 2021) \cite{ravi2021cholan} &  &  &  &  &  &  &  & \checkmark & &-\\ \hline
    \textbf{Zhang et al.} (DASFAA 2021) \cite{zhang2021attention} &  &  & description &  &  &  &  & \checkmark & &- \\ \hline

  \end{tabular}}
  \vspace{-6mm}
\end{table*}

\vspace{-3.5mm}
\section{Embedding}
\label{embeddings}
\vspace{-1mm}

The core idea of embedding is to represent the meaning of a piece of natural language by using a low-dimensional real-valued dense vector. Based on the given documents and KBs, diverse embedding techniques could be leveraged to capture the linguistic patterns and common sense knowledge in text, such as semantic roles, syntactic structures, and lexical meanings. In this section, we introduce four kinds of embeddings used in DL based EL methods: (1) word embedding; (2) mention embedding; (3) entity embedding; (4) alignment embedding. Considering the strong correlation between context embedding and context similarity feature, we will introduce the context embedding in Section \ref{sec:context embedding}.

\vspace{-2.5mm}
\subsection{Word Embedding}
\label{word embedding}

A word embedding can map words from a vocabulary to vectors of real numbers to represent their meaning. In this subsection, we classify word embeddings used by most EL works into two categories: (1) pre-trained; (2) learned.

\subsubsection{Pre-trained}
\label{pre-trained}

Pre-trained language models can learn widely applicable embeddings of words based on their co-occurrences and neighborhoods in a large quantity of text corpora. They learn word embeddings in advance and store them in lookup tables. Lots of DL based EL methods directly use a pre-trained (fixed) lookup table to get word embeddings. Specifically, EL methods \cite{zwicklbauer2016robust,ganea2017deep,kolitsas2018end,yang2018collective,yang2019learning,le2019boosting,hou2020improving,chen2020improving,hu2020graph} utilized a Word2Vec \cite{mikolov2013efficient} lookup table to get word embeddings. Additionally, several DL based EL methods \cite{gupta2017entity,le2018improving,le2019distant,le2019boosting,gillick2019learning,fang2019joint,moon2018multimodal} exploited a GloVe \cite{pennington2014glove} lookup table as the word embedding source.

\vspace{-1.5mm}
\subsubsection{Learned}
\label{learned}

Pre-trained embedding may not be suitable for domain-specific data sets which contain string tokens with highly specialized semantics. In this case, many DL based EL methods learn a domain-specific word embedding using some embedding techniques. 
DL based EL methods \cite{cao2017bridge,cao2018neural,yamada2016joint,sun2015modeling,van2020rel,sil2017neural,francis2016capturing,phan2018pair,phan2017neupl,eshel2017named,xue2019neural,mueller2018effective,nguyen2016joint,fang2016entity} learned word embeddings via Word2Vec based on the huge corpora such as Wikipedia. Word2Vec contains continuous bag-of-words (CBOW) model and skip-gram (SG) model \cite{mikolov2013efficient}. In reality, EL methods \cite{yamada2016joint,cao2017bridge,cao2018neural,sun2015modeling,van2020rel,phan2018pair,phan2017neupl,eshel2017named,mueller2018effective} prefer to use SG model for training. 
Additionally, Shahbazi et al. \cite{shahbazi2019entity} utilized ELMo \cite{peters2018deep} to learn word embeddings. ELMo is a large-scale context-sensitive language model that uses bi-directional long short-term memory (Bi-LSTM) \cite{lample2016neural} in forward and backward directions to encode word embeddings depending on its context. Adjali et al. \cite{adjali2020ecir} used Sent2Vec \cite{pagliardini2018unsupervised} which extends the CBOW model to learn representations of words.

Some word embedding aims to obtain numeric representations of words by leveraging their character-level information. The main idea is that words are made of morphemes, i.e., meaningful sequences of characters with different lengths. Injecting character-level information into the final word embedding could effectively handle the issue of out-of-vocabulary (OOV) words \cite{kim2016character}. For example, OOV words often occur due to misspellings, and substrings of the word can be leveraged to approximate its embedding. 
Mueller and Durrett \cite{mueller2018effective} and Onoe and Durrett \cite{onoe2020fine} utilized CNNs \cite{lecun1998gradient} to generate character-level representations of words concatenated with the learned word embedding, which makes their models recognize character-level correspondences between entity mention and candidate entity better. Kolitsas et al. \cite{kolitsas2018end} and Chen et al. \cite{chen2021lightweight} utilized Bi-LSTM to capture significant character lexical information. Both of them generated the character-dependent embedding of a word by concatenating the forward hidden state of its last character, the backward hidden state of its first character, and its pre-trained word embedding.

Generally, word embedding cannot represent the order information of words. Intuitively, the position-dependent signal can reinforce semantics by incorporating the order information. Some EL systems add the positional embedding to the word embedding to explicitly encode the relative/absolute positions of words as vectors. 
Sun et al. \cite{sun2015modeling} and Le and Titov \cite{le2019distant} leveraged a positional embedding that was modeled by the distance between the word and the entity mention in a given piece of text. Xue et al. \cite{xue2019neural} followed Vaswani et al. \cite{vaswani2017attention} to define the positional embedding of each word in the entity mention. Specifically, they took the position index of each word as input and utilized a pre-defined sinusoidal function of the position index as its positional encoding.

\vspace{-2.5mm}
\subsection{Mention Embedding}
\label{mention embedding}

Mention embedding is a learned representation for an entity mention. It is usually used to model the similarity between the entity mention and the candidate entity. For some EL methods \cite{sun2015modeling,wu2020dynamic}, it is simply obtained by averaging the embeddings of words which compose the entity mention.

Additionally, several DL based EL models utilize CNNs to generate vectors of entity mentions. Specifically, each word in the entity mention is encoded into a word embedding using a pre-trained Word2Vec lookup table, and thus a sequence of vectors $\mathbf{w}_1, \mathbf{w}_2, ..., \mathbf{w}_n$ can be found, where $n$ is the word length of the entity mention. Francis-Landau et al. \cite{francis2016capturing} mapped the vector sequence into a fixed-size vector using a CNN parameterized with a filter bank $\mathcal{W} \in \mathbb{R}^{k \times dz}$, where $k$ is the number of filter maps, $d$ is the dimension of words, and $z$ is the width of the convolution. Then they put the result through a \textit{ReLU} function and combined the results with sum-pooling, based on the following formulation:

\begin{equation}
\vspace{-0.4mm}
  conv(w_{1:n}) = \sum_{i=1}^{n-z}\max\{0, \mathcal{W}\mathbf{w}_{i:i+z}\} 
  \vspace{-0.1mm}
\end{equation}

\noindent where $\mathbf{w}_{i:i+z}$ is a concatenation of word embeddings and the $\max$ is element-wise. Some other EL methods \cite{nguyen2016joint,yang2019learning,xue2019neural,chen2021lightweight} also utilized this similar process to obtain the mention embedding.

Moreover, due to the high ambiguity of entity mentions, Cao et al. \cite{cao2017bridge,cao2018neural} proposed an embedding model, which can learn multiple sense embeddings for each entity mention to denote its different meanings. Specifically, each entity mention is first mapped to a set of shared mention senses according to a pre-defined dictionary. To train the mention sense embedding, they combined each mention sense and its context to predict the corresponding named entity by extending the CBOW model. 

Kolitsas et al. \cite{kolitsas2018end} defined a ``soft head'' embedding to capture key components of the entity mention, which is built using an attention mechanism on top of the entity mention's word embeddings. They concatenated the embeddings of the first, last, and the ``soft head'' words of the entity mention, and then leveraged a shallow feedforward neural network (FFNN) to produce the representation $\mathbf{m}$ of the entity mention as follows:

\begin{equation}
\vspace{-0.4mm}
  \mathbf{m} = \mathrm{FFNN}([\mathbf{m}_f; \mathbf{m}_l; \mathbf{m}_s])
  \vspace{-0.4mm}
\end{equation}

\noindent where $\mathbf{m}_f$, $\mathbf{m}_l$, $\mathbf{m}_s$ are the representations of the first, last, and ``soft head'' words respectively. 

Wu et al. \cite{wu2020dynamic} took the entity mention with the text where it appears as input and leveraged ELMo \cite{peters2018deep} to learn a contextualized representation for the entity mention. What's more, Onoe and Durrett \cite{onoe2020fine} fed the representation of the entity mention learned by ELMo into a Bi-LSTM encoder followed by a span attention layer to get the final mention embedding.

Li et al. \cite{li2020efficient} and Yamada et al. \cite{yamada2020global} leveraged the pre-trained language model BERT \cite{devlin2019bert} to encode the entity mention. When training the BERT model, the masked language model and the next sentence prediction model are trained together, with the goal of minimizing the combined loss function of the two strategies. The masked language model randomly masks some tokens in text, and then independently recovers these masked tokens by conditioning on the encoding vectors obtained via a bi-directional Transformer. The next sentence prediction model receives pairs of sentences as input and learns to predict if the second sentence in the pair is the subsequent sentence in the original document. Li et al. \cite{li2020efficient} took the entity mention with its context as a sequence of words as input to BERT and the outputs of tokens of the last layer were selected as the contextualized embeddings of words. They averaged the output embeddings of the entity mention words to generate the mention embedding. Yamada et al. \cite{yamada2020global} leveraged BERT to encode an input sequence consisting of words in the document and masked entity tokens corresponding to the entity mention. For each entity mention, taking the BERT output vector of its corresponding masked entity as input, the mention embedding is learned via a single-layer perceptron.

\vspace{-2.5mm}
\subsection{Entity Embedding}
\label{sec:entity embedding}

Entity embedding, which represents each entity in a KB in a continuous vector space, is important for DL based EL. It is helpful for capturing the entity-entity topical coherence feature (described in Section \ref{sec:topicalcoherence}) by mapping similar entities close to each other in the same space. Entity embedding is also used as an important raw material to calculate the context similarity feature, which will be introduced in Section \ref{sec:contextsimilarity}. KBs like Wikipedia provide lots of structured and textual data for learning entity embeddings, such as surface form, entity description, entity context, and type information. Some EL methods leverage one type of them, while others combine several types to learn the entity embedding.

\vspace{-1.5mm}
\subsubsection{Surface Form}
\label{sec:surface form}

The most direct way to get entity embeddings is using entity surface forms to learn lexical representations. Surface form consists of a word or a sequence of words like ``Paris'' or ``New York City''. It is usually used as the matching sequence to locate the corresponding candidate entity due to its uniqueness in the reference KB \cite{charton2014improving}. Sun et al. \cite{sun2015modeling} obtained the entity surface form embedding by averaging the embeddings of words which compose the entity surface form based on learned word embeddings. Several EL methods \cite{francis2016capturing,nguyen2016joint,yang2019learning,xue2019neural} used CNNs to convert each entity surface form into a vector, which is the same as the way of using CNNs to generate mention embeddings introduced in Section \ref{mention embedding}. What's more, Chen et al. \cite{chen2021lightweight} ran a Bi-LSTM on the entity surface form to get a representation for each entity.

\vspace{-1.5mm}
\subsubsection{Entity Description}
\label{sec:description}
Entity description is a piece of text that introduces useful background information about the entity. It usually provides a concise summary of salient properties for the entity. For instance, in the description page of the entity ``Los Angeles Lakers'' in Wikipedia, much useful information such as players, championships, and history can be found. A great quantity of EL works consider it as a good resource during the process of constructing entity embeddings, and take the representation of the entity description as the important auxiliary information to reinforce the semantic information of entities. It is also noted that zero-shot setting has been a surge in the field of EL \cite{logeswaran2019zero,wu2020scalable,li2020efficient,tang2021bidirectional} where each named entity is defined only by its entity description, while other information such as type information and relations between entities are absent. Thus, it is essential to design a suitable method to learn the representation of the entity description to obtain the entity embedding.

He et al. \cite{he2013learning} built a stacked denoising auto-encoders (DA) \cite{vincent2008extracting} to encode the entity description. Stacked DA is able to capture general concepts of the input and ignore noise. It takes a one-hot vector of the entity description as input and outputs a learned representation. 

Huang et al. \cite{huang2015leveraging} first represented each entity description as a bag-of-words, which is then transformed by a word hashing layer into letter tri-gram vectors. They applied a deep neural network (DNN) on these tri-gram vectors to learn useful semantic entity embedding based on the entity description. 

Sil et al. \cite{sil2017neural} computed a weighted average of embeddings of all words in the entity description by using the inverse document frequency of each word as the weight. They further applied a fully connected \textit{tanh} activation layer to this obtained embedding to get the final entity description embedding.  

Lots of studies \cite{francis2016capturing,nguyen2016joint,yang2019learning,gupta2017entity,xue2019neural} utilized CNNs to distill the entity description into a meaningful topic embedding. They combined this embedding with the entity surface form embedding mentioned in Section \ref{sec:surface form} to form the final entity embedding. 

Doc2Vec \cite{le2014distributed} is a modification of Word2Vec and can learn fixed-size embeddings from variable-length pieces of text like entity descriptions. Both Zwicklbauer et al. \cite{zwicklbauer2016robust} and Sevgili et al. \cite{sevgili2019improving} leveraged Doc2Vec to generate an entity embedding based on the entity description.

Phan et al. \cite{phan2017neupl} and Fang et al. \cite{fang2019joint} both used a single-directional LSTM network to encode the entity description. Instead of taking the last hidden state as the representation for entity description, Phan et al. \cite{phan2017neupl} applied max-pooling over all hidden state vectors to produce a fixed-size entity embedding.

Ganea and Hofmann \cite{ganea2017deep} collected co-occurrence words from the entity description of each entity and regarded them as positive words of that entity for learning its embedding. They leveraged the word-entity co-occurrence counts to define a practical approximation of a word-entity conditional distribution. In addition, they sampled negative words unrelated to that entity from a generic word distribution. Finally, they used a max-margin loss to infer the optimal embedding of the entity with a goal that vectors of positive words are closer to the embedding of that entity compared with vectors of negative words. Many EL methods \cite{kolitsas2018end,le2018improving,yang2018collective,fang2019joint,yang2019learning,le2019boosting,wu2020dynamic} follow this work and use the same method to generate the entity embedding.

Hu et al. \cite{hu2020graph} extracted words with the highest TF-IDF from the entity description as related words of the entity. To learn better representations of entities by aggregating the semantic information from their neighboring entities and words, they constructed a heterogeneous entity-word graph for each document, which consists of entity nodes corresponding to the candidate entities of mentions in the document, word nodes corresponding to the related words of candidate entities, and relationships added by computing the similarity score between nodes based on pre-trained word embeddings. They applied Graph Convolutional Network (GCN) \cite{kipf2016semi} on this graph to encode the global semantic information among candidate entities and finally generated entity embeddings.

Sometimes, entity descriptions are a little too long to accurately express semantic information of entities. To alleviate the long-term dependency problem and extract worthy information, some DL based EL systems leverage BERT \cite{devlin2019bert} to encode the entity description in order to learn the entity embedding. Specifically, several EL works \cite{wu2020scalable,li2020efficient,fang2020high,zhang2021attention} took the entity description as a sequence of words as input to BERT. Most of these works \cite{wu2020scalable,li2020efficient,zhang2021attention} inserted a special start token $\mathtt{[CLS]}$ at the beginning of the input sequence and the output of the last layer at this start token produced by the Transformer encoder is regarded as the vector representation of the input sequence. Fang et al. \cite{fang2020high} obtained the entity embedding via average-pooling over the hidden states of all description tokens in the last BERT layer.

However, BERT can only take a limited length (i.e., $512$ tokens) of the entity description as input. To tackle this problem, Tang et al. \cite{tang2021bidirectional} proposed a multi-paragraph reading model for discovering more textual information in the entity description which is composed of multiple paragraphs. Specifically, they first utilized BERT to encode the concatenation of the mention context and the entity description paragraph to obtain a representation for each paragraph. These representations are then used as the input of a multi-head attention module proposed in Transformer \cite{vaswani2017attention} to gather the semantic dependence among the paragraphs. Finally, a weighted-pooling layer is applied on hidden states output by the multi-head attention module to produce the entity embedding.

\vspace{-1.5mm}
\subsubsection{Entity Context}
\label{knowledge}

In addition to entity descriptions, KBs like Wikipedia can provide additional valuable context information for named entities. Some DL based EL methods utilize various kinds of entity context to learn entity embeddings, such as anchor texts, fact information, and neighbor entities. 

Anchor text in Wikipedia is a hyperlink from an entity mention in an entity page linking to its corresponding entity page, and can help the learning of entity embeddings in two aspects. First, the anchor text is a link that jumps from one entity page to another, providing rich entity-entity co-occurrence information. What's more, anchor texts residing in documents supply abundant context words for their referring entities, and thus they can offer rich word-entity co-occurrence information.

Specifically, Yamada et al. \cite{yamada2016joint} used anchor texts in Wikipedia to learn the entity embedding based on the entity-entity co-occurrence information they provide. Inspired by Wikipedia Link-based Measure (WLM) \cite{milne2008learning}, which is a standard entity relatedness measure in traditional EL, they assumed that entities with similar hyperlinks are related. For example, the teams ``Boston Celtics'' and ``Toronto Raptors'' are highly related because they have many common hyperlinks to the entity pages about ``NBA''. Their model simply learned to place entities with similar hyperlinks near one another in the vector space via maximizing the conditional probability $P(e_o|e)$, where $e_o$ is one of the hyperlinks of the entity $e$, which is similar to the idea of the SG model in Word2Vec \cite{mikolov2013efficient}.
Fang et al. \cite{fang2016entity} exploited the entity-entity co-occurrence information offered by anchor texts to learn the entity embedding by placing the representations of two entities connected by anchor texts close to each other in the vector space.
Sevgili et al. \cite{sevgili2019improving} used anchor texts in Wikipedia to create a graph whose nodes are entities and edges are the hyperlinks between entities. Then the vector representation of each entity is generated by running DeepWalk \cite{perozzi2014deepwalk} over the constructed graph.

Starting from Ganea and Hofmann \cite{ganea2017deep}, many EL methods \cite{kolitsas2018end,le2018improving,yang2018collective,fang2019joint,yang2019learning,le2019boosting,wu2020dynamic,hu2020graph} took word-entity co-occurrence counts provided by anchor texts as another source to learn entity embeddings. We have introduced their specific learning method in Section \ref{sec:description}.
Additionally, lots of EL approaches \cite{yamada2016joint,fang2016entity,cao2017bridge,phan2017neupl,eshel2017named,mueller2018effective,wu2020dynamic,van2020rel,phan2018pair} utilized anchor texts to construct entity embeddings by mapping embeddings of words and entities into the same vector space. In these approaches, similar words and entities are placed close to each other in a vector space. More details will be introduced in Section \ref{alignment}.

Recently, Yamada et al. \cite{yamada2020global} utilized anchor texts residing in Wikipedia to learn the contextualized entity embedding. Specifically, they proposed a new masked entity prediction model to learn the entity embedding, inspired by the masked language model adopted in BERT \cite{devlin2019bert}. Masked entity prediction aims to predict randomly masked entities based on words and non-masked entities. They trained this model based on the contexts of the referring entities provided by the anchor texts in Wikipedia.

A KB is usually composed of facts about entities. Facts are also important information for modeling entity embeddings. We denote a fact as $\langle h, r, o\rangle$, where $h$ is a head entity, $r$ is a relation, and $o$ is a tail entity. 
Fang et al. \cite{fang2016entity} followed TransE \cite{bordes2013translating} to define a score function $s(\cdot)$ for a fact $\langle h, r, o\rangle$ based on embeddings of $h$, $r$, and $o$ as follows:

\vspace{-0.6mm}
\begin{equation}
\vspace{-0.6mm}
\label{fact}
  s(h, r, o) = b - \frac{1}{2}{\parallel \mathbf{h}+\mathbf{r}-\mathbf{o}\parallel}^2
  \vspace{-0.4mm}
\end{equation}

\noindent where $\mathbf{h}$, $\mathbf{r}$, $\mathbf{o}$ are the embeddings of $h$, $r$, $o$ respectively, and $b$ is a constant for numerical stability. After maximizing the above score function, the representations of $h$ and $o$ can be used as entity embeddings.
Moon et al. \cite{moon2018multimodal} also leveraged facts to generate entity embeddings as follows:

\vspace{-1mm}
\begin{equation}
\vspace{-1mm}
  s(h, r, o) = score(\mathbf{h}, \mathbf{r}, \mathbf{o})
  \vspace{-1mm}
\end{equation}
  \vspace{-1mm}

\noindent where $score()$ is a deep neural network that produces a likelihood of a valid fact $\langle h, r, o\rangle$.
For a certain entity $e$, Huang et al. \cite{huang2015leveraging} defined a connected entity set $\mathcal{E}(e)$ containing its corresponding head entities and tail entities existing in the same facts, which is another form of entity-entity co-occurrence information. For each entity in $\mathcal{E}(e)$, they obtained its surface form and represented it as a bag-of-words. The vectors of all connected entities are concatenated to generate the final entity embedding. 
Cao et al. \cite{cao2017bridge,cao2018neural} extended the SG model in Word2Vec by maximizing the log probability of observing its connected entity set given an entity as follows:

\vspace{-1mm}
\begin{equation}
\label{sg}
  \mathcal{L} = \sum_{e }^{}\log P(\mathcal{E}(e)|e) 
\end{equation}
\vspace{-2mm}

\noindent Thus, entities sharing many common connected entities tend to have similar representations.

Zwicklbauer et al. \cite{zwicklbauer2016robust} defined the neighbor entities of entity $e$ as entities that appear surrounding $e$ in text. They generated a corpus that exclusively comprises entities sequentially by replacing all available linked surface forms in documents with its corresponding target entities and removing all non-entity identifiers like words and punctuations. Similar to Equation \ref{sg}, they leveraged the SG model in Word2Vec over this created corpus to learn entity embeddings by leveraging the neighbor entities in a window.

\vspace{-1.5mm}
\subsubsection{Type Information}
\label{sec:typeinformation}

A KB consists of a type hierarchy and entities that are instances of types. The entity type is a kind of important information to represent the semantics of entities. A large number of EL methods have leveraged entity type information to learn entity embeddings.

Sun et al. \cite{sun2015modeling} averaged embeddings of entity type words based on the learned word embeddings to get the entity type embedding. They utilized a low-rank neural tensor network \cite{socher2013recursive} to jointly encode the entity type embedding and the entity surface form embedding introduced in Section \ref{sec:surface form} to learn the final entity representation. 
Gupta et al. \cite{gupta2017entity} calculated a relevant probability $P(t|e)$ of observing type $t$ given entity $e$ as $\mathrm{sigmoid}(\mathbf{t}  \cdot \mathbf{e})$, where $\mathbf{t}$ and $\mathbf{e}$ are type embedding and entity embedding respectively. They maximized this probability to learn entity and type embeddings jointly and injected type embeddings into entity embeddings ultimately.

Huang et al. \cite{huang2015leveraging} extracted a set of attached entity types for each entity based on Freebase \cite{bollacker2008freebase}, represented each entity type as a one-hot vector, and used it as one of the inputs for DNN to learn entity embeddings. 
Wu et al. \cite{wu2020dynamic} extracted the notable type from Freebase for each entity and utilized the average embedding of words in the type string as a type embedding. They then obtained the final entity embedding by concatenating the type embedding with the learned entity embedding \cite{ganea2017deep}. 
Le and Titov \cite{le2019distant} only used type information to produce the entity embedding. They first leveraged Freebase to get a type set for each entity. Each type $t$ of the entity $e$ is assigned a vector $\mathbf{t}$ based on the pre-trained word embeddings, and then the representation $\mathbf{e}$ of the entity can be calculated as follows:

\begin{equation}
  \mathbf{e} = ReLU(\mathcal{W}\frac{1}{|T_e|}\sum_{t \in T_e}^{}\mathbf{t} + b)
\end{equation}

\noindent where $T_e$ is the type set of the entity $e$, $\mathcal{W}$ is a weight matrix and $b$ is a bias vector.

Chen et al. \cite{chen2020improving} proposed to inject latent type information into the entity embedding via modeling the immediate context where the entity appears. They considered that context consistency is a strong proxy for type compatibility, so they leveraged the pre-trained BERT to encode entity contexts that are randomly sampled from Wikipedia for that entity. The representation of the entity is computed by aggregating all the context representations via average-pooling.

Hou et al. \cite{hou2020improving} exploited the embeddings of semantic types to generate entity embeddings. They first created a dictionary of fine-grained semantic types, and then extracted semantic types from its Wikipedia article for each entity. To construct semantic reinforced entity embedding, they averaged the semantic type embeddings obtained from Word2Vec and combined them with learned entity embeddings \cite{ganea2017deep} via linear aggregation. 

\vspace{-2.5mm}
\subsection{Alignment Embedding}
\label{alignment}

As introduced in Sections \ref{word embedding} and \ref{sec:entity embedding}, the word embedding and the entity embedding could be learned in different ways based on various data. However, in many cases, these two categories of embeddings are learned separately and are not in the same vector space. This makes it impossible to calculate the similarity between words and entities effectively, which is an essential operation for computing the context similarity feature in EL. Therefore, aligning word and entity embeddings into the same vector space (called alignment embedding) is very necessary and important so that similar words and entities are placed close to each other in a common space. After alignment, we can measure the similarity between words and entities by simply computing the cosine similarity of their embeddings. We introduce some classic alignment embedding techniques as follows.

As mentioned in Section \ref{knowledge}, anchor text is a key resource for aligning embeddings of words and entities. The window words surrounding the anchor text could be regarded as context words of its referring entity, and could provide ample word-entity co-occurrence examples, which could be leveraged to align embeddings.
Yamada et al. \cite{yamada2016joint} proposed an alignment model which leverages anchor texts and their context words by extending the SG model in Word2Vec. The objective function is defined to predict the surrounding context words of the referring entity of the anchor text: 

\begin{equation}
\vspace{-0.5mm}
\mathcal{L} =  \sum_{(e, Q)\in A}^{}\sum_{w\in Q}^{}\log P(w|e)  
\vspace{-0.1mm}
\end{equation}

\noindent where $A$ denotes the set of anchor texts, $e$ denotes the referring entity of the anchor text, $Q$ is the set of its surrounding context words, and $w$ is a word in $Q$. 

Moreover, based on the word-entity co-occurrence counts collected from anchor texts and entity descriptions, Fang et al. \cite{fang2016entity} maximized the score function $s(\cdot)$ to shorten the distance between each co-occurrence pair:

\begin{equation}
\label{eq:distance}
  s(w, e) = b - \frac{1}{2} {\parallel \mathbf{w} - \mathbf{e} \parallel}^2
\end{equation}

\noindent where $\mathbf{w}$ and $\mathbf{e}$ represent the corresponding embeddings for word $w$ and entity $e$ respectively, and $b$ is a constant.

Additionally, Eshel et al. \cite{eshel2017named} used Word2Vecf embedding algorithm \cite{levy2014dependency} to train word and entity embeddings jointly via leveraging the word-entity co-occurrence counts collected from the set of frequent words appearing in the Wikipedia article of that entity.

\vspace{-2.5mm}
\section{Feature}
\label{features}

Features are designed to calculate the similarity between the entity mention and the candidate entity in various aspects. Benefiting from DL, more and more EL methods have utilized embeddings as features directly or used different embeddings as a source for features design. In addition, some traditional features which are manually designed, such as the prior popularity feature, are still applied in recent DL based EL methods. In this section, we review five categories of features found to be effective and broadly used to rank candidate entities in DL based EL methods, i.e., prior popularity, surface form similarity, type similarity, context similarity, and topical coherence.

\vspace{-2.5mm}
\subsection{Prior Popularity}
\label{sec:priorpopilarity}

The prior popularity is the probability of the appearance of a candidate entity given an entity mention without considering the context where the mention appears. It is a simple but strong feature, and almost all DL based EL methods leverage this feature. For example, the former U.S. president ``Barack Obama'' is more popular than his wife ``Michelle Obama'', both of which could be referred to by ``Obama''. In most cases when people mention ``Obama'', they mean the former president rather than his wife. Anchor text in Wikipedia is the most broadly used source to estimate the prior popularity feature. Given an entity mention $m$, the prior popularity feature $p(e|m)$ of a candidate entity $e$ can be estimated as follows:

\begin{equation}
  p(e|m) = \frac{count(m, e)}{count(m)} 
\end{equation}

\noindent 
where $count(m)$ denotes the number of anchor texts having the entity mention $m$ as the surface form in Wikipedia; $count(m, e)$ represents the number of anchor texts with the surface form $m$ pointing to the candidate entity $e$. In many common real-world entity linking data sets, the accuracy of using this prior popularity feature alone can reach $70\%-85\%$ \cite{ganea2017deep}, which demonstrates its effectiveness in EL.

\vspace{-2.5mm}
\subsection{Surface Form Similarity}

Intuitively, an entity mention and a candidate entity with the same or similar surface forms are likely to indicate a gold mapping. For example, the entity mention ``Univ Manchester'' is more likely to refer to the university ``The University of Manchester'' rather than the football team ``Manchester City F.C.'' due to their greater surface form similarity. A great many DL based EL systems leverage some traditional surface form similarity measures, such as edit distance, Dice coefficient score, character Dice, skip bigram Dice, and left and right Hamming distance scores. We have introduced them in our previous survey \cite{shen2014entity} and would not introduce them in detail here.

In addition, several DL based EL methods \cite{francis2016capturing,nguyen2016joint,yang2019learning,xue2019neural} considered the cosine similarity between the mention embedding introduced in Section \ref{mention embedding} and the entity surface form embedding introduced in Section \ref{sec:surface form} as the surface form similarity feature. Chen et al. \cite{chen2021lightweight} leveraged CNNs over the mention embedding and the entity surface form embedding to extract n-gram features of the entity mention and the candidate entity respectively. A two-layer fully connected neural network is then applied to the concatenation of the output embeddings to obtain the surface form similarity feature. What's more, Moon et al. \cite{moon2018multimodal} trained a separate deep neural network to encode surface forms of the entity mention and the candidate entity, which produces a purely lexical embedding without semantic allusion. The surface form similarity feature between the entity mention and the candidate entity is computed based on their distance in the embedding space. 

\vspace{-2.5mm}
\subsection{Type Similarity}

In addition to being used as auxiliary information to construct entity embeddings introduced in Section \ref{sec:typeinformation}, type information is utilized by several DL based EL methods to calculate the type similarity feature. For example, knowing that the correct type of the entity mention ``Boston'' in some context is \textit{sports\_team} could constrain its corresponding entity with relevant types, such as sport or team. Generally, KBs contain rich type information for a candidate entity, while the type information of an entity mention in some context is absent. Some typing systems are proposed to predict types of an entity mention with the help of its surrounding context.

Raiman and Raiman \cite{raiman2018deeptype} proposed DeepType, which is a method to solve EL only using type constraints. They restricted the types in their typing system via selecting a set of parent-child relations over the ontology in Wikipedia. Then they leveraged a Bi-LSTM classifier to obtain the type conditional probability $P(t|m,c)$ for type $t$ given the entity mention $m$ and its surrounding context $c$. For a candidate entity $e$ belonging to types $t_1, ..., t_n$, the type similarity feature $f_{typ}(m, e)$ is calculated as follows:

\vspace{-1mm}
\begin{equation}
\begin{split}
  f_{typ}(m, e) = 1-\beta +\beta \cdot\prod_{i=1}^{n}(1- \alpha _i+\alpha _i\cdot P(t_i|m,c))
\end{split}
\end{equation}

\noindent where $\beta$ is a smoothing parameter over all types and $\alpha_i$ is a smoothing parameter per type.

Onoe and Durrett \cite{onoe2020fine} proposed a fine-grained typing system, which contains tens of thousands of types derived from Wikipedia categories. They selected a single linear layer to decode the concatenation of the mention embedding introduced in Section \ref{mention embedding} and the context embedding which will be introduced in Section \ref{sec:context embedding}. They then utilized a \textit{sigmoid} function to normalize the output vector of the decoder and obtained the type conditional probability $P(t|m,c)$ for type $t$ given the entity mention $m$ and its surrounding context $c$. The type similarity feature $f_{typ}(m, e)$ between the entity mention $m$ and the candidate entity $e$ of types $t_1, ..., t_n$ is calculated as follows:

\vspace{-0.5mm}
\begin{equation}
  f_{typ}(m, e) = \sum_{i=1}^{n} P(t_i|m,c)
\end{equation}

Yang et al. \cite{yang2019learning} trained a typing system proposed by Xu and Barbosa \cite{xu2018neural}. They concatenated the mention embedding and the context embedding to form a representation, which is then used as the input of a softmax classifier to predict the probability distribution over four types (i.e., PER, GPE, ORG, and UNK) of the entity mention. The type similarity feature is measured as the similarity between the predicted type distribution of the entity mention and the types of the candidate entity.

\vspace{-2.5mm}
\subsection{Context Similarity}
\label{sec:contextsimilarity}

The most straightforward feature for entity linking is to measure the similarity between the representation derived from the context around the entity mention and the representation associated with the candidate entity. To model this context similarity feature, the corresponding context of the entity mention would be encoded as a context embedding using various neural architectures, and then diverse computing methods would be utilized to calculate the similarity score between the generated context embedding and the entity embedding mentioned in Section \ref{sec:entity embedding}. In the following, we first describe how to generate the context embedding, and subsequently show the computing methods for calculating this context similarity feature.

\vspace{-1.5mm}
\subsubsection{Context Embedding}
\label{sec:context embedding}

With the advent of latent embeddings, traditional bag-of-words \cite{guo2013link,zhang2010nus,vstajner2009entity} and keyphrase \cite{hoffart2011robust,hoffart2012kore} models may seem to be superseded by neural models, which can implicitly capture semantic and syntactic information of the context. Formally, given an entity mention $m$ in a sentence $(w_j^{\mathrm{left}}, ..., w_1^{\mathrm{left}}, m, w_1^{\mathrm{right}}, ..., w_j^{\mathrm{right}})$ within a pre-defined window size $j$, the context of $m$ contains the left-side $j$ context words of $m$ as $(w_j^{\mathrm{left}}, ..., w_1^{\mathrm{left}})$ and the right-side $j$ context words of $m$ as $(w_1^{\mathrm{right}}, ..., w_j^{\mathrm{right}})$. Context embedding aims to represent the surrounding context of $m$ as a dense and low-dimensional vector. As the context is composed of words, a large number of DL based EL methods leverage word embeddings introduced in Section \ref{word embedding} as input to learn the context embedding based on various neural models. For the simplest case, some EL methods \cite{yamada2016joint,cao2017bridge,yang2018collective,shahbazi2019entity,adjali2020ecir} averaged the embeddings of context words to derive the context embedding. For the other cases, we group models of generating context embeddings into several categories based on their different neural architectures, such as RNNs, CNNs, attention, Transformers, and others. In the remainder, we will describe these categories of context embedding models in detail.

\paratitle{RNNs-based.} RNN is designed to process data that is sequential in nature, especially when the input sequence has variable-length. Accordingly, the RNNs-based neural models are appropriate for context learning. RNN intends to capture word dependencies and context structures using the recurrent unit, which is an NN that is shared between all time steps \cite{mudgal2018deep}. At time step $l$, the recurrent unit takes the $l$-th input $\mathbf{x}_l$ and the hidden state vector of the previous time step $\mathbf{h}_{l-1}$ to produce the hidden state vector of the current time step $\mathbf{h}_l$. Hence, $\mathbf{h}_l$ contains the previous and current input information $\mathbf{x}_1, ..., \mathbf{x}_l$.

LSTM \cite{hochreiter1997long} is the most popular RNN architecture, which is designed to better maintain the long term memory. It introduces a memory cell to remember values over arbitrary time intervals, and uses three kinds of gates (input gate, output gate, and forget gate) to regulate the flow of information into and out of the cell. Fang et al. \cite{fang2019joint} utilized a single-directional LSTM to encode the context as an embedding, while some works \cite{le2019distant,onoe2020fine,wu2020dynamic,chen2021lightweight} used a Bi-LSTM \cite{lample2016neural} to encode context. Additionally, a few DL based EL methods \cite{phan2017neupl,gupta2017entity,sil2017neural} used a forward LSTM and a backward LSTM to encode the left-side and right-side context of the entity mention respectively. Specifically, Phan et al. \cite{phan2017neupl} leveraged the hidden state vectors to generate the context embedding, while the others \cite{gupta2017entity,sil2017neural} used the output vectors. It is noted that these two LSTMs are different from Bi-LSTM since each LSTM of them only embeds half of the context while each LSTM of Bi-LSTM encodes the full context.

A slight variation of LSTM is the Gated Recurrent Unit (GRU) \cite{cho2014learning}. Compared with LSTM, it is simpler by combining the input and forget gate into a single update gate, and has been adopted by some DL based EL methods such as \cite{mueller2018effective,eshel2017named}. Both of them utilized a forward GRU and a backward GRU to encode the left-side and right-side context respectively. The outputs of the forward GRU and the backward GRU are leveraged to generate the context embedding.

\paratitle{CNNs-based.} RNN is trained to recognize patterns across time, while CNN learns to recognize patterns across space \cite{lecun1998gradient}. RNN works well when the long-term semantics is required, while CNN works well when detecting local and position-invariant patterns is important, which might be a keyphrase that expresses a particular sentiment \cite{minaee2020deep}. Thus, contexts in CNNs-based models are not supposed to be too long. CNN consists of multiple convolutional layers, each of which extracts local features around each context word, and the size of the output of the convolutional layers depends on the number of words in context. The context embedding is constructed via combining local feature vectors extracted by convolutional layers.

Several DL based EL methods \cite{sun2015modeling,nguyen2016joint,francis2016capturing,xue2019neural} applied CNNs over the context words to generate the hidden vector sequences, which were then transformed by a non-linear function and pooled by sum-pooling \cite{nguyen2016joint,francis2016capturing,xue2019neural} or average-pooling \cite{sun2015modeling} to generate the context embedding. 

\begin{figure}[t]
  \centering
  \includegraphics[width=0.5\textwidth]{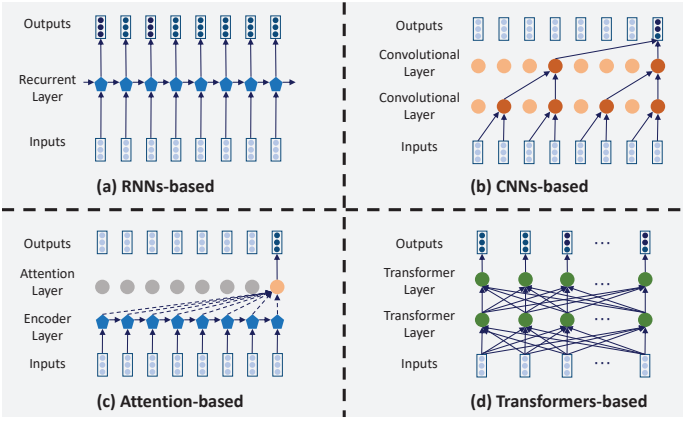}
  \caption{Architectures of different neural networks for learning the context embedding.}
  \label{fig:rnncnnattention}
  \vspace{-6mm}
\end{figure}

\paratitle{Attention-based.} Starting from Bahdanau et al. \cite{bahdanau2015neural}, attention mechanism has become an increasingly popular concept and useful tool in many NLP tasks. The idea of attention is to highlight the informative parts of the input based on an attention vector. In order to give different attention weights to the context words in learning the context embedding, attention-based EL models estimate how strongly they are correlated with the attention vector. Then these models take the weighted sum of the values of the context words as the context embedding. To illustrate more clearly, we propose a general formula to introduce various attention-based models for learning the context embedding, which is generally defined as a sum of the context word values weighted by the correlation between the value and the attention vector:

\vspace{-2.5mm}
\begin{equation}
  \mathbf{c} = F(U(\mathcal{V},\mathcal{A}))\mathcal{V}
\end{equation}

\noindent where $\mathcal{V}$ and $\mathcal{A}$ correspond to the matrices of values and attention vectors respectively, $U(\cdot)$ denotes a function used to calculate the attention weight of each value, $F(\cdot)$ usually denotes a \textit{softmax} function, and $\mathbf{c}$ is the context embedding. We summary some prominent attention-based EL models by introducing different choices of values, attention vectors, and $U(\cdot)$ as follows.

In learning the context embedding, values are usually components of the context. Lots of attention-based EL models \cite{ganea2017deep,cao2018neural,kolitsas2018end,hou2020improving,yang2019learning,le2018improving,le2019boosting,hu2020graph} used embeddings of context words as values. Phan et al. \cite{phan2017neupl} took hidden state vectors of the forward LSTM and the backward LSTM as values, while Wu et al. \cite{wu2020dynamic} and Chen et al. \cite{chen2021lightweight} took outputs of the Bi-LSTM as values. Eshel et al. \cite{eshel2017named} and Mueller and Durrett \cite{mueller2018effective} regarded outputs of the forward GRU and the backward GRU as values.

Attention vectors are used to provide background information and discover the important and relevant parts of values. Some EL works \cite{phan2017neupl,mueller2018effective,eshel2017named,cao2018neural,wu2020dynamic} chose the same attention vector for all values, which is the embedding of the candidate entity. Several DL based EL models \cite{ganea2017deep,kolitsas2018end,hou2020improving,yang2019learning,le2018improving,le2019boosting,hu2020graph} assumed that a context word is important if it is strongly related to at least one candidate entity, so they selected the embedding of the most related candidate entity for each context word as the attention vector. Additionally, Zhang et al. \cite{zhang2021attention} regarded the representations of connected entities of the candidate entity introduced in Section \ref{knowledge} as the attention vectors.

$U(\cdot)$ is a function used to determine the attention weight of each value by measuring the similarity or correlation between the value and the attention vector. Cao et al. \cite{cao2018neural} used the cosine similarity, Mueller and Durrett \cite{mueller2018effective} and Zhang et al. \cite{zhang2021attention} leveraged the dot product, and a great many attention-based models \cite{ganea2017deep,kolitsas2018end,hou2020improving,yang2019learning,le2018improving,le2019boosting,wu2020dynamic,hu2020graph} utilized the bilinear similarity. We will introduce these three metrics in detail in Section \ref{sec:similaritycomputation} as they are also utilized to calculate the context similarity feature. Additionally, Phan et al. \cite{phan2017neupl} and Eshel et al. \cite{eshel2017named} obtained the similarity by leveraging a single-layer perceptron to encode the value and the attention vector.

\paratitle{Transformers-based.} RNNs-based models suffer from the sequential processing of the context, which is one of their computation bottlenecks. Despite that CNNs-based models are less sequential, the computational cost to capture relations between words in the context also grows with the increasing length of the context. Transformers \cite{vaswani2017attention} overcome this limitation via applying self-attention mechanism to calculate an attention weight for each word in the context, in order to model the influence each word has on another \cite{minaee2020deep}. Recently, Transformers-based pre-trained language models use much deeper neural architectures compared with models based on RNNs and CNNs, and are pre-trained on much larger text corpora to learn contextualized text representations by predicting words conditioned on their surrounding contexts. These pre-trained language models could be fine-tuned using task-specific labels of downstream tasks like EL.

BERT \cite{devlin2019bert} is a widely used Transformers-based pre-trained language model in context embedding learning for EL task. Several DL based EL methods \cite{wu2020scalable,fang2020high,zhang2021attention} took the word-pieces of the entity mention and its context as the input of BERT. Wu et al. \cite{wu2020scalable} and Zhang et al. \cite{zhang2021attention} inserted a start token $\mathtt{[CLS]}$ and regarded the output of the last layer at this start token as the context embedding, while Fang et al. \cite{fang2020high} obtained the context embedding via average-pooling over the hidden states of all context tokens in the last layer. Tang et al. \cite{tang2021bidirectional} leveraged the multi-paragraph reading model based on BERT introduced in Section \ref{sec:description} to encode the context as a representation. Compared with the model introduced in Section \ref{sec:description}, they added a new multi-head attention module which leverages the learned entity embedding as the query to emphasize the importance of the context which is encoded in the entity embedding.

Architectures of RNNs-based, CNNs-based, attention-based, and Transformers-based models for learning the context embedding are shown in Figure \ref{fig:rnncnnattention}.

\paratitle{Other Methods.} There are also some other methods to learn context embeddings, which we briefly introduce as follows. He et al. \cite{he2013learning} used stacked DA \cite{vincent2008extracting} to encode the context as a representation, while Zwicklbauer et al. \cite{zwicklbauer2016robust} and Sevgili et al. \cite{sevgili2019improving} utilized Doc2Vec \cite{le2014distributed} to generate an embedding for the context, which is the same way for them to learn the entity embedding based on the entity description introduced in Section \ref{sec:description}.

\vspace{-1.5mm}
\subsubsection{Context Similarity Feature Computation}
\label{sec:similaritycomputation}

After the generation of the context embedding described above, we could compute the context similarity feature based on some similarity measures between the context embedding of the entity mention and the entity embedding of the candidate entity. Here, we overview the context similarity measures used by DL based EL models in the following. 

Cosine similarity is a commonly used similarity measure for real-valued vectors. Based on it, the context similarity feature $f_{cnt}(m,e)$ between the entity mention $m$ and the candidate entity $e$ can be defined as follows:

\begin{equation}
\label{cosine}
  f_{cnt}(m,e) = \frac{\mathbf{c}\cdot \mathbf{e}}{\parallel \mathbf{c} \parallel \parallel \mathbf{e} \parallel } 
\end{equation}

\noindent where $\mathbf{c}$ is the context embedding for the entity mention $m$, $\mathbf{e}$ is the entity embedding of the candidate entity $e$, and $\parallel \mathbf{c} \parallel$ and $\parallel \mathbf{e} \parallel$ are lengths of two embeddings respectively. Cosine is a trigonometric function that helps to describe the orientation of two vectors. The highest similarity value $1$ is reserved for the two vectors that are the most close together, while the lowest similarity value $0$ is reserved for the two vectors that are the least close together. Lots of DL based EL methods \cite{sun2015modeling,yamada2016joint,francis2016capturing,zwicklbauer2016robust,nguyen2016joint,cao2017bridge,yang2018collective,cao2018neural,xue2019neural,chen2020improving,wu2020dynamic,van2020rel,phan2018pair,sil2017neural,gillick2019learning,chen2021lightweight} applied this metric to calculate the context similarity feature.

Additionally, the numerator of Equation \ref{cosine}, which is the dot product of the context embedding $\mathbf{c}$ and the entity embedding $\mathbf{e}$, has been directly used to measure the context similarity feature by several EL methods \cite{he2013learning,kolitsas2018end,wu2020scalable,li2020efficient,yamada2020global,moon2018multimodal}. Compared with cosine similarity that cares about angle difference between two vectors, dot product focuses on both angle and magnitude.

Starting from Ganea and Hofmann \cite{ganea2017deep}, bilinear similarity has been utilized by several DL based EL systems \cite{le2018improving,yang2019learning,le2019boosting,wu2020dynamic,hou2020improving,van2020rel,hu2020graph} to compute the context similarity feature, which can be denoted as follows:

\vspace{-1.5mm}
\begin{equation}
  f_{cnt}(m,e) = \mathbf{c}\mathbf{\mathcal{W}}\mathbf{e}^\mathrm{T}
\end{equation}

\noindent where $\mathbf{c}$ is the context embedding for the entity mention $m$, $\mathbf{e}$ is the entity embedding of the candidate entity $e$, and $\mathcal{W}$ is a trainable diagonal matrix, which indicates the relationships between two embeddings.

Moreover, several EL works utilized neural architectures to calculate the context similarity feature. Specifically, some DL based EL methods \cite{mueller2018effective,le2019distant,fang2020high} passed the concatenation of the context embedding and the candidate entity embedding to a single-layer perceptron to learn a context similarity score for each candidate entity, while Sevgili et al. \cite{sevgili2019improving} and Fang et al. \cite{fang2019joint} used an MLP instead. Some DL based EL methods \cite{logeswaran2019zero,gu2021read,ravi2021cholan} concatenated the mention context and the entity description as a sequence pair together with a special start token $\mathtt{[CLS]}$ and separator tokens $\mathtt{[SEP]}$ as the input of BERT, while Wu et al. \cite{wu2020scalable} concatenated the context embedding described in Section \ref{sec:context embedding} and the entity embedding described in Section \ref{sec:description} as the input of BERT. All these works regarded the output of the last hidden layer at the start token as the representation of the input pair. Some of these works \cite{logeswaran2019zero,wu2020scalable,ravi2021cholan} applied a linear layer to this representation to produce the context similarity feature, while Gu et al. \cite{gu2021read} utilized an MLP with softmax function over this representation.

\vspace{-2.5mm}
\subsection{Topical Coherence}
\label{sec:topicalcoherence}

The aforementioned features mainly concern the local similarity between the entity mention and one of its candidate entities, and each entity mention in the document could be linked independently based on these features, which are often called local features. Starting from Cucerzan et al. \cite{cucerzan2007large}, more and more EL methods take into consideration the global topical coherence among the referring entities within the same document, where entity mentions are linked collectively. It is based on an assumption that co-occurring entity mentions in the same document often refer to topically coherent entities. Given a document $D$, the topical coherence feature $f_{coh}(e|D)$ for a candidate entity $e$ of an entity mention $m$ could be calculated via averaging the topical coherence between the candidate entity $e$ and assigned mapping entities of an entity mention set in the same document:

\vspace{-1.5mm}
\begin{equation}
\label{eq:topicalcoherence}
   f_{coh}(e|D) = \frac{1}{|M_s|} \sum_{m^{\prime} \in M_s}^{} \psi (e, e_{m^{\prime}}^*)
 \end{equation}
 \vspace{-1.5mm} 

\noindent where $M_s$ is a mention set in the document $D$, $e_{m^{\prime}}^*$ is the assigned mapping entity of the entity mention $m^{\prime}$ in the mention set $M_s$, and $\psi(\cdot)$ is a function to calculate a pairwise similarity between entities which may indicate the degree of their topical coherence. In order to introduce the various methods of calculating the topical coherence feature clearly, we summarize the process of this feature generation in the following three steps: (1) mention set selection; (2) mapping entity assignment; and (3) topical coherence feature computation. Firstly, a mention set $M_{s}$ whose mapping entities are the targets that the candidate entity $e$ needs to be coherent with is selected from all the entity mentions in the document $D$. Next, for each entity mention $m^{\prime}$ in the mention set $M_s$, we \textit{temporarily} assign a mapping entity $e_{m^{\prime}}^{*}$ to it from its candidate entities when computing this feature, as its real mapping entity is unknown to us and needs to be figured out in this entity linking task. Finally, the topical coherence feature is obtained by computing the average similarity between the candidate entity $e$ and assigned mapping entities $e_{m^{\prime}}^{*}$. An illustration of this process is shown in Figure 4.

\begin{figure}[t]
  \centering
  \includegraphics[width=0.44\textwidth]{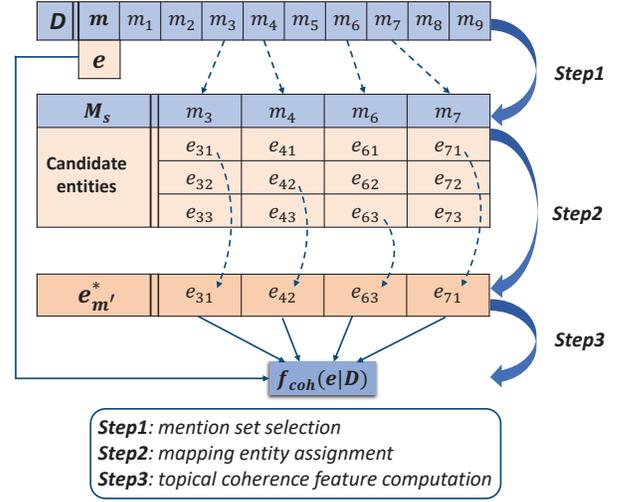}
  \caption{The process of calculating the topical coherence feature $f_{coh}(e|D)$ for a candidate entity $e$ of an entity mention $m$ in a document $D$.}
  \label{fig:topical}
  \vspace{-6mm}
\end{figure}

The key component in the third step of feature computation is the similarity function $\psi(\cdot)$ for a pair of entities. Entity embeddings introduced in Section \ref{sec:entity embedding} are usually utilized in this step to calculate the entity similarity. Specifically, a large number of DL based EL methods \cite{zwicklbauer2016robust,phan2017neupl,kolitsas2018end,xue2019neural,cao2017bridge,zwicklbauer2016robust,huang2015leveraging,cao2018neural,yang2018collective,nguyen2016joint,phan2018pair} leveraged cosine similarity as the similarity function and several EL methods \cite{yamada2016joint,ganea2017deep,le2018improving,le2019boosting,yang2019learning,chen2020improving,van2020rel,hou2020improving,wu2020dynamic,hu2020graph} exploited bilinear similarity to calculate the entity similarity based on entity embeddings. 
Another well-known method of entity similarity is WLM \cite{milne2008learning}, which is a powerful but simple method based on anchor texts in Wikipedia. More details of WLM can be found in our previous survey \cite{shen2014entity}. Some DL based EL methods \cite{yang2018collective,xue2019neural,phan2018pair} used WLM as the similarity function $\psi(\cdot)$.
Moreover, starting from Le and Titov \cite{le2018improving}, some EL works \cite{le2019boosting,hou2020improving,van2020rel} incorporated relations between entities as latent variables to compute pairwise entity similarity. 
Fang et al. \cite{fang2016entity} utilized a distance based metric which is similar to Equation \ref{eq:distance}. 
Some hand-crafted similarity measurements such as the number of KB relations existing between entities \cite{globerson2016collective} and entity-entity co-occurrence counts obtained from Wikipedia \cite{yang2018collective} were also leveraged to calculate the entity similarity.

However, even if we get the pairwise similarity function between entities, it is still hard to compute the topical coherence feature for the candidate entity. According to some works \cite{yamada2016joint,ganea2017deep,yang2019learning,le2019boosting}, the optimization of this feature is an NP-hard problem since the second step (i.e., mapping entity assignment) that needs to assign a mapping entity for each entity mention in $M_{s}$, is impossible to be implemented before the entity linking task is finished as its real mapping entity is unknown to us. To solve this problem, several DL based EL methods obtain the assigned mapping entity for an entity mention temporarily by selecting an entity from its candidate entities, which are called simplified methods. In addition, some DL based EL methods do not obtain the assigned mapping entities explicitly but leverage the whole set of candidate entities to compute the topical coherence feature, which are called optimized methods. In the remainder of this subsection, we introduce simplified methods and optimized methods respectively.

\vspace{-1.5mm}
\subsubsection{Simplified Methods}
\label{simplifiedmethods}

Simplified methods try to transform the NP-hard problem into a computable problem by determining a \textit{temporary} mapping entity for each entity mention in the mention set in the calculation of the topical coherence feature. Here, we mainly focus on how to determine a temporary mapping entity for an entity mention. According to the range of entity mentions that are selected to form the mention set in the first step (i.e., mention set selection), we classify the simplified methods into the following two groups: (1) whole coherence; (2) partial coherence.

\paratitle{Whole coherence.}  
The whole coherence means all the other entity mentions in the document $D$ are selected to form the mention set $M_s$ except the entity mention $m$ of the candidate entity $e$ for which we calculate the topical coherence feature via Equation \ref{eq:topicalcoherence}.

Some EL method \cite{shen2012linden} chose the candidate entity with the highest prior popularity feature as the assigned mapping entity $e_{m^{\prime}}^*$ for the entity mention $m^{\prime}$ in $M_s$, which is defined as follows:

\vspace{-1.5mm}
\begin{equation}
\label{eq:topical_prior}
\vspace{-0.5mm}
  e_{m^{\prime}}^* = \mathop{\arg\max}\limits_{e^{\prime}\in C(m^{\prime})} P(e^{\prime}|m^{\prime})
  \vspace{-0.1mm}
\end{equation}
\vspace{-1.5mm}

\noindent where $C(m^{\prime})$ is a set of candidate entities for entity mention $m^{\prime}$ and $P(e^{\prime}|m^{\prime})$ is the prior popularity feature introduced in Section \ref{sec:priorpopilarity}. Additionally, several DL based EL methods \cite{globerson2016collective,le2019boosting,fang2016entity,wu2020dynamic} regarded the most similar candidate entity to the candidate entity $e$ in Equation \ref{eq:topicalcoherence} as the assigned mapping entity $e_{m^{\prime}}^*$ for the entity mention $m^{\prime}$ in $M_s$, which is defined as follows:

\vspace{-1.5mm}
\begin{equation}
\label{eq:topical_max}
\vspace{-0.5mm}
  e_{m^{\prime}}^* = \mathop{\arg\max}\limits_{e^{\prime}\in C(m^{\prime})} \psi (e, e^{\prime})
  \vspace{-0.1mm}
\end{equation}
\vspace{-1.5mm}

\noindent where $\psi(\cdot)$ is an entity similarity function as we have introduced earlier. 

\paratitle{Partial coherence.} 
Different from the whole coherence, the partial coherence means that the candidate entity $e$ in Equation \ref{eq:topicalcoherence} only needs to be coherent with the assigned mapping entities of parts of entity mentions in the same document, which significantly reduces computational complexity and time consumption.

Yamada et al. \cite{yamada2016joint} regarded a set of unambiguous entity mentions in the document $D$ as the mention set $M_s$ and considered an entity mention unambiguous when the prior popularity feature of one of its candidate entities is greater than $0.95$. They used Equation \ref{eq:topical_prior} to obtain the assigned mapping entity $e_{m^{\prime}}^*$ for the entity mention $m^{\prime}$ in $M_s$. The topical coherence feature for the candidate entity is computed as the similarity between the candidate entity embedding and the averaged embedding of assigned mapping entities of unambiguous entity mentions.

Kolitsas et al. \cite{kolitsas2018end} leveraged the candidate entities having high local scores derived from local features as assigned mapping entities $e_{m^{\prime}}^{*}$ to participate in the calculation of the topical coherence feature. The entity mentions of these selected candidate entities form the mention set $M_s$. To obtain the topical coherence feature for each candidate entity, they summed up the representations of assigned mapping entities and calculated the similarity between the candidate entity embedding and this generated representation. 

It is noted that in the above two simplified methods, candidate entities of different entity mentions have a fixed mention set throughout the EL process, while in the following simplified methods, the mention set changes with candidate entities of different entity mentions.

Fang et al. \cite{fang2016entity} and Chen et al. \cite{chen2021lightweight} selected entity mentions around the entity mention $m$ of the candidate entity $e$ in Equation \ref{eq:topicalcoherence} in a pre-defined window as the mention set $M_s$. They obtained the topical coherence feature for the candidate entity by averaging the similarities between the candidate entity embedding and embeddings of assigned mapping entities defined via Equation \ref{eq:topical_max}.

A great number of DL based EL methods link entity mentions in a document in a sequential manner and utilize already disambiguated entities to help generate the topical coherence feature for the subsequent candidate entities. When we calculate this feature for some candidate entity $e$ via Equation \ref{eq:topicalcoherence}, the mention set $M_s$ is composed of entity mentions which have already been linked before the entity mention $m$ of the candidate entity $e$ and those already disambiguated entities are considered as assigned mapping entities. Specifically, Zwicklbauer et al. \cite{zwicklbauer2016robust} created a topic vector, which is set by summing up the representations of already disambiguated entities. Once the linking of some mention is completed, the topic vector changes accordingly. The topical coherence feature for the candidate entity is defined as the similarity between the candidate entity embedding and the topic vector.
Nguyen et al. \cite{nguyen2016joint} used GRUs to encode already disambiguated entities with an assumption that the hidden state vector of GRUs could summarize the information about the already disambiguated entities based on the characteristics of RNN. The topical coherence feature for the candidate entity is calculated as the similarity between the candidate entity embedding and the current hidden state vector of GRUs. 
Similarly, Fang et al. \cite{fang2019joint} utilized an LSTM as the encoder instead. 
What's more, to generate a topical coherence feature vector for each candidate entity, Yang et al. \cite{yang2018collective} first calculated the similarities between the candidate entity embedding and the embeddings of already disambiguated entities, and then concatenated the maximal and average similarity score to form the topical coherence feature vector. 
Some EL methods \cite{cao2017bridge,fang2019joint,yamada2020global,gu2021read} used a simple to complex strategy, i.e., began to link entity mentions that are easier to be disambiguated and then utilized information provided by already disambiguated entities to generate topical coherence features for subsequent entity mentions that are relatively more difficult to be disambiguated. 
Yang et al. \cite{yang2019learning} proposed the dynamic context augmentation (DCA), whose basic idea is to accumulate knowledge from already disambiguated entities as a dynamic context to enhance later decisions. They applied an attention mechanism on DCA to adjust the weights of already disambiguated entities. The topical coherence feature for the candidate entity is obtained by summing up the similarity scores between the candidate entity embedding and the embeddings of already disambiguated entities. Chen et al. \cite{chen2020improving} also applied DCA in their global model.
Gu et al. \cite{gu2021read} utilized already disambiguated entities to enhance the later decisions through a dynamic multi-turn way. Specifically, once an entity mention $m_{i-1}$ is linked, the context for identifying the entity mention $m_i$ will be updated by replacing $m_{i-1}$ with its assigned mapping entity  $e_{m-1}^*$. Then the updated context is leveraged to generate the topical coherence feature for $m_i$ using the BERT \cite{devlin2019bert} encoder in each turn. To alleviate error propagation caused by falsely linked entities, a gate mechanism is introduced on the current and historical representations to control which part of history cues should be inherited.

In addition, Phan et al. \cite{phan2017neupl,phan2018pair} proposed pair-linking, which is based on an assumption that each candidate entity only needs to be coherent with another candidate entity. They selected the most compatible candidate entity for the candidate entity $e$ in Equation \ref{eq:topicalcoherence} from candidate entities of all entity mentions in the document $D$ as the assigned mapping entity. The topical coherence feature of the candidate entity $e$ could be computed based on local features of the pair of the candidate entity and the assigned mapping entity as well as the similarity between them.
\vspace{-1.5mm}
\subsubsection{Optimized Methods}
\label{optimizedmethods}

Different from simplified methods, optimized methods do not select a temporary mapping entity but keep all candidate entities for the entity mention in the mention set $M_s$ when computing the topical coherence feature. Thus, optimizing the topical coherence objective is intractable due to the large number of candidate entities. Optimized methods usually leverage graph-based approaches to perform optimization as they can dynamically capture the global interdependence between different candidate entities in the document and learn the topical coherence feature for each candidate entity. We will introduce some optimized methods in detail as follows. 

We first introduce three EL methods based on Graph Neural Network (GNN) to automatically decide the relevant candidate entities of entity mentions in the mention set and then generate a topical coherence feature for each candidate entity. Cao et al. \cite{cao2018neural} applied GCN \cite{kipf2016semi} whose basic idea is to enhance the feature of a node according to its neighbor nodes to integrate topical coherence information. The entity mentions around the entity mention $m$ in a pre-defined window are regarded as the mention set $M_s$. To obtain the topical coherence feature for each candidate entity, they constructed a graph for the candidate entity by taking this candidate entity and all candidate entities of entity mentions in the mention set $M_s$ as nodes and relations between them extracted from a KB as edges. Then they applied a GCN on this graph to aggregate information from linked nodes to the candidate entity node. Given the hidden state $\mathbf{h}_{l-1}$ of the ($l-1$)-th layer, graph convolution is operated to generate the hidden state $\mathbf{h}_l$ as follows:

\begin{equation}
    \mathbf{h}_l = f(\mathcal{W}_l(\tilde{A}\mathbf{h}_{l-1}) + b_l)
\end{equation}

\noindent where $\tilde{A}$ is a normalized adjacent matrix of the input graph with self-connection, $\mathcal{W}_l$ and $b_l$ are the weights and bias in the $l$-th layer respectively, and $f(\cdot)$ represents a non-linear activation function. After $\mathcal{T}$ times of the graph convolution, the hidden state $\mathbf{h}_\mathcal{T}$ integrates information from both the candidate entity and its neighbor nodes, and it is treated as a topical coherence feature of the candidate entity.

Similarly, Wu et al. \cite{wu2020dynamic} leveraged a dynamic GCN learning paradigm to calculate the topical coherence feature. The graph structure of $\tilde{A}$ is dynamically computed and modified by a graph weight determinator during training in each layer, and it can capture topical coherence better than GCN with a fixed graph structure. Fang et al. \cite{fang2020high} utilized a Graph Attention Network (GAT) model to encode the global topical coherence information of the candidate entity. It is noted that GAT is a typical dynamic GCN, which can dynamically change nodes and edges according to the current state. Moreover, with the help of the masked self-attention layers, GAT can implicitly assign different importance to different neighbor nodes and capture the relevance between candidate entities well.

In addition, Ganea and Hofmann \cite{ganea2017deep} formulated this problem based on a binary Conditional Random Field (CRF) model by taking candidate entities of all entity mentions in the document as nodes. Considering the exact maximum a posteriori inference on this CRF is NP-hard, they adopted loopy belief propagation (LBP) as an approximate inference method based on $\mathcal{T}$ times message-passing iterations to produce a marginal probability for each candidate entity as its topical coherence feature. This optimized method is followed by several DL based EL methods \cite{le2018improving,le2019boosting,chen2020improving,van2020rel,hou2020improving,hu2020graph}.

\vspace{-2.5mm}
\section{Algorithm}
\label{algorithm}

Algorithm takes features introduced in the above section as input and outputs the final entity linking result. Specifically, algorithm aims to select the target mapping entity for each entity mention based on the local and global features of its candidate entities. Traditional EL methods pursued some algorithms to select the mapping entity, such as linear model \cite{ratinov2011local,shen2012linden,shen2012liege}, support vector machines \cite{chen2011collaborative,pilz2011names}, logistic classifier \cite{fang2016entity,monahan2011cross} and Na\"ive Bayes classifier \cite{varma2009iiit}. Recently, DL based EL methods that we focus on in this survey used algorithms, such as MLP, PageRank, graph regularization, and reinforcement learning. In this section, we introduce such algorithms by summarizing them into three groups, namely, MLP, graph-based algorithms, and RL. 

\vspace{-2.5mm}
\subsection{MLP}

For each candidate entity $e$ of the entity mention $m$, MLP could transform its local and global features to a ranking score $\Phi(m,e)$ via a learned non-linear transformation. We have introduced its basic structure in Section \ref{whydl}. Some DL based EL methods \cite{wu2020dynamic,le2019distant,cao2018neural,le2019boosting,eshel2017named,sil2017neural} leveraged a single-layer perceptron to encode features to produce a ranking score for the candidate entity. For each entity mention, the candidate entity with the highest ranking score is chosen as the output mapping entity. Additionally, several EL works \cite{van2020rel,chen2020improving,hou2020improving,sevgili2019improving,fang2020high,yang2019learning,le2018improving,kolitsas2018end,gillick2019learning,mueller2018effective,ganea2017deep,shahbazi2019entity,tang2021bidirectional,chen2021lightweight,hu2020graph,adjali2020ecir} leveraged an MLP instead. 

MLP could be trained using the stochastic gradient descent algorithm and require a loss function to guide the parameter learning. Loss functions are used to evaluate how well the specific algorithm models the given data. If the predicted mapping results output by the algorithm deviate too much from the gold mapping results, loss functions would generate large values. Thus, loss functions need to be minimized during the training of MLP. There are two widely used loss functions which we introduce briefly.

A large quantity of MLP methods \cite{fang2020high,van2020rel,chen2020improving,wu2020dynamic,hou2020improving,le2019distant,yang2019learning,le2018improving,kolitsas2018end,ganea2017deep,le2019boosting,chen2021lightweight,hu2020graph} leveraged the max-margin loss, which tries to make the ranking score of the gold mapping entity higher than the ranking scores of the other candidate entities by a safety margin. The max-margin loss of a training instance is defined as follows:

\vspace{-2mm}
\begin{equation}
\vspace{-0.5mm}
   \mathcal{L} = \max (0, \gamma - \Phi (m, e^+) + \Phi (m, e^-))
   \vspace{-0.2mm}
 \end{equation} 

\noindent where $\Phi (m, e)$ denotes the ranking score of the candidate entity $e$ given the entity mention $m$, and $\gamma$ is the safety margin. Each training instance is constructed by a positive gold mapping entity $e^+$ with a negative entity $e^-$, with the purpose to make the ranking score of $e^+$ be at least a margin $\gamma$ larger than that of $e^-$.

Cross-entropy loss is another loss function utilized by some MLP methods \cite{cao2018neural,eshel2017named,gillick2019learning,tang2021bidirectional,adjali2020ecir}. The binary cross-entropy loss which increases when the predicted label diverges from the actual label is defined as follows:

\vspace{-3mm}
\begin{equation}
\vspace{-0.5mm}
  \mathcal{L} = -(y\log \Phi (m, e) + (1-y) \log ({1-\Phi (m, e)}))
\vspace{-0.2 mm}
\end{equation}

\noindent where $y \in \{0, 1\}$ denotes the actual label of the candidate entity. If the candidate entity $e$ is the gold mapping entity for the entity mention $m$, the value of $y$ is $1$; otherwise $0$. $\Phi (m, e) \in (0, 1)$ indicates the ranking score of the candidate entity $e$ given the entity mention $m$. 

\vspace{-2.5mm}
\subsection{Graph-Based Algorithms}

In general, graph-based algorithms are usually leveraged by collective EL methods, which make decisions on mapping entities jointly for all entity mentions in the same document. Specifically, for each document containing a set of entity mentions, the graph-based algorithm first needs to construct a graph by taking the candidate entities of all entity mentions in the document as nodes and similarity scores between candidate entities as edge weights. Next, a graph-based ranking algorithm is performed on this graph to assign a ranking score to each candidate entity, which represents its degree of importance in the overall graph structure. Finally, for each entity mention, the candidate entity with the highest ranking score is chosen as the output mapping entity. In the following, we introduce graph-based algorithms leveraged by DL based EL methods in detail.

Zwicklbauer et al. \cite{zwicklbauer2016robust} applied PageRank on a graph which is composed of the candidate entities of all entity mentions in the same document and a topic vector node introduced in Section \ref{simplifiedmethods}. The transition matrix of the graph describes the likelihood of walking from a node to the adjacent node and is calculated as the harmonic mean between two nodes' local scores. They employed the prior popularity feature introduced in Section \ref{sec:priorpopilarity} as a jump probability for each candidate entity node. Ultimately, they applied the PageRank algorithm over the constructed graph to compute a ranking score for each candidate entity.

Xue et al. \cite{xue2019neural} introduced recurrent random-walk layers for collective EL, which reinforce the evidence for related EL decisions into high probability decisions with the help of external KB. The graph constructed for each document contains the candidate entities of all entity mentions in the document. To define the transition matrix between candidate entities, they calculated a relevance score for each pair of candidate entities by summing up their semantic similarity score and the WLM \cite{milne2008learning} score based on Wikipedia. By introducing $\mathcal{T}$ random-walk layers, they can easily propagate evidence for $\mathcal{T}$ times and produce a ranking score for each candidate entity.

\begin{table*}[t]
  \centering
  \caption{List of ten widely used data sets for entity linking. ``\# \textbf{M.}'' refers to the number of entity mentions, ``\# \textbf{D.}'' refers to the number of documents, and ``\# \textbf{M./D.}'' refers to the number of entity mentions per document in the data set.}  
  \label{tb:datasets}
  \begin{tabular}{c|c|c|c|c|c|c|c}
    \hline
    \textbf{Data set (Abbreviation)} & \textbf{Genre} & \textbf{Year} & \textbf{KB} & \# \textbf{M.} & \# \textbf{D.} & \# \textbf{M./D.}  & \textbf{URL} \\ \hline
    \textbf{MSNBC} \cite{cucerzan2007large} & news & 2007 & Wikipedia & 656 & 20 & 32.80 & \href{https://cogcomp.seas.upenn.edu/page/resource_view/4}{\tabincell{c}{https://cogcomp.seas.upenn.edu/page\\/resource\_view/4}} \\\hline
    \textbf{AQUAINT (AQ)}  \cite{milne2008learning} & news & 2008 & Wikipedia & 449 & 50 & 8.98 & \href{http://community.nzdl.org/wikification/docs.html}{\tabincell{c}{http://community.nzdl.org/wikification\\/docs.html}}\\\hline
    \textbf{TAC-KBP2010 (KBP)}  \cite{ji2011knowledge} & news, blogs & 2010 & Wikipedia & 3750 & 3684 & 1.02 & \href{https://tac.nist.gov/2010/KBP/data.html}{https://tac.nist.gov/2010/KBP/data.html} \\\hline
    \textbf{AIDA-CoNLL (AIDA)}  \cite{hoffart2011robust} & news & 2011 & \tabincell{c}{YAGO/Freebase/\\Wikipedia} & 34587 & 1393 & 24.83 & \href{http://resources.mpi-inf.mpg.de/yago-naga/aida/download/}{\tabincell{c}{http://resources.mpi-inf.mpg.de/yago-naga\\/aida/download/}} \\ \hline
    \textbf{ACE2004 (ACE)}  \cite{ratinov2011local} & news & 2011 & Wikipedia & 257 & 35 & 7.34 & \href{https://cogcomp.seas.upenn.edu/page/resource_view/4}{\tabincell{c}{https://cogcomp.seas.upenn.edu/page\\/resource\_view/4}}  \\\hline
    \textbf{KORE50 (KORE)}  \cite{hoffart2012kore} & tweets & 2012 & YAGO/DBpedia & 148 & 50 & 2.96 & \href{http://resources.mpi-inf.mpg.de/yago-naga/aida/download/}{\tabincell{c}{http://resources.mpi-inf.mpg.de/yago-naga\\/aida/download/}}  \\\hline
    \textbf{N3-RSS500 (RSS)}  \cite{roder2014n3} & RSS-feeds & 2014 & DBpedia & 1000 & 500 & 2.00 & \href{http://aksw.org/Projects/N3NERNEDNIF.html}{http://aksw.org/Projects/N3NERNEDNIF.html}  \\\hline
    \textbf{N3-Reuters128 (Reuters)}  \cite{roder2014n3} & news & 2014 & DBpedia & 881 & 128 & 6.88 & \href{http://aksw.org/Projects/N3NERNEDNIF.html}{http://aksw.org/Projects/N3NERNEDNIF.html} \\ \hline
    \textbf{WNED-CWEB (CW)}  \cite{guo2018robust} & news & 2016 & Wikipedia & 11154 & 320 & 34.86 & \href{https://doi.org/10.7939/DVN/10968}{https://doi.org/10.7939/DVN/10968} \\\hline
    \textbf{WNED-WIKI (WI)}  \cite{guo2018robust} & news & 2016 & Wikipedia & 6821 & 320 & 21.32 & \href{https://doi.org/10.7939/DVN/10968}{https://doi.org/10.7939/DVN/10968} \\
    \hline
  \end{tabular}
\end{table*}

What's more, Huang et al. \cite{huang2015leveraging} leveraged graph regularization to perform collective inference. They constructed a graph for each document and each node in the graph represents a pair of entity mention and one of its candidate entities. A weighted edge is added between two nodes if satisfying some constraints and the weight is computed as the semantic similarity between two candidate entities. They first initialized a ranking score for each candidate entity based on the linear combination of local features. Some nodes with high prior popularity feature introduced in Section \ref{sec:priorpopilarity} are regarded as labeled seed nodes, which remain unchanged during the graph regularization. Then graph regularization is applied on this graph to refine the ranking scores of unlabeled nodes by means of the labeled seed nodes, with an assumption that two strongly connected nodes should have similar ranking scores.

\vspace{-2.5mm}
\subsection{Reinforcement Learning}

Reinforcement learning (RL) is an area of ML concerned with how software agents ought to perform discrete actions in an environment according to a policy, which is trained to maximize some cumulative rewards \cite{kaelbling1996reinforcement}. Using RL as a ranking algorithm for entity linking, a set of candidate entities are selected by agents as the entity mapping results which can maximize the sum of expected rewards.

Fang et al. \cite{fang2019joint} and Yang et al. \cite{yang2019learning} regarded the EL task as a sequence decision problem and leveraged RL algorithm to obtain the entity mapping results. In their models, the agent is designed as a policy network which can learn a stochastic policy and prevent the agent from getting stuck at an intermediate state. Under the guidance of the policy network, the agent decides which action (i.e., choosing the target mapping entity from candidate entities) should be taken at each state (i.e., current local and global encoding). After performing all decisions in the episode, each action will get an expected reward and the goal is to maximize the total expected rewards $\mathcal{J}$, which is defined as follows:

\vspace{-2mm}
\begin{equation}
\vspace{-2.5mm}
\begin{split}
  \mathcal{J} = \sum_{l}^{} \sum_{a}^{} &\pi (a|st)R(a_l)\\
  R(a_l) = p(a_l)\sum_{i=l}^{L}p(a_i)+(&1-p(a_l))(\sum_{i=l}^{L} p(a_i) +l-L ) 
\end{split}  
\vspace{-1.5mm}
\end{equation}

\noindent where $\pi (a|st)$ is the policy network indicating the probability of taking the action $a$ under the state $st$, and $R(a_l)$ is the expected reward of the action $a$ at $l$-th time step. To compute the expected reward $R(a_l)$, $p(a_l) \in \{1, 0\}$ indicates whether the current action is correct or not, and $\sum_{i=l}^{L}p(a_i)$ and $-(\sum_{i=l}^{L} p(a_i) +l-L) $ represent the number of correct actions and that of wrong ones from time $l$ to the end of episode respectively. Accordingly, RL could explore the long-term influence of current selection on subsequent decisions.

\vspace{-2.5mm}
\section{Data Sets and Evaluation}
\label{evaluation}

In this section, we first introduce several widely used real-world EL data sets, tools, and evaluation metrics. Then we give a quantitative performance analysis of representative DL based EL methods.

\vspace{-2.5mm}
\subsection{Data Sets and Tools}
\label{sec:data sets}

Lots of data sets with different properties (e.g., genre, year, KB, and the number of entity mentions per document) have been used to evaluate EL systems. Table \ref{tb:datasets} shows an overview of ten well-known public EL data sets used by DL based EL methods. We introduce these data sets in detail as follows:

\begin{itemize}
  \item \textbf{MSNBC} \cite{cucerzan2007large} is annotated from MSNBC news articles and contains documents from $10$ different domains (i.e., two documents per domain).
  \item \textbf{AQUAINT} \cite{milne2008learning} contains documents collected from the Xinhua News Service, New York Times, and the Associated Press. Each document contains about $250$ to $300$ words, where the first entity mention of an entity is manually annotated to Wikipedia.
  \item \textbf{TAC-KBP2010} \cite{ji2011knowledge} contains news and blogs from various agencies. It is constructed for the TAC $2010$ conference and only contains approximately one entity mention per document which is unsuitable to model the topical coherence feature. 
  \item \textbf{AIDA-CoNLL} \cite{hoffart2011robust} is an annotated corpus of Reuters news documents. It contains much more documents than other existing EL data sets and is manually linked to KBs by authors. To train and test EL methods, the data set is often divided into three parts: AIDA-Train for training, AIDA-A for validation, and AIDA-B for testing.
  \item \textbf{ACE2004} \cite{ratinov2011local} is a subset of ACE2004 \cite{doddington2004automatic} coreference documents annotated using Amazon Mechanical Turk\footnote{https://www.mturk.com/}.
  \item \textbf{KORE50} \cite{hoffart2012kore} is extracted from some microblogging platform (i.e., Twitter). It contains short documents (i.e., tweets) on various domains (e.g., music, business, sports, and celebrities). Each tweet consists of a few sentences with some ambiguous entity mentions, and most entity mentions are first names referring to persons with high level of ambiguity.
  \item \textbf{N3-RSS500} \cite{roder2014n3} is created using a data set of RSS-feeds (i.e., short formal documents), which are from major international newspapers. The data set covers a wide range of domains, such as world, business, and science. 
  \item \textbf{N3-Reuters128} \cite{roder2014n3} contains economic news documents extracted from the Reuters-21587 corpus\footnote{http://kdd.ics.uci.edu/databases/reuters21578/reuters21578.html}. Both N3-RSS500 and N3-Reuters128 are manually annotated by R{\"o}der et al. \cite{roder2014n3}.
  \item \textbf{WNED-CWEB} \cite{guo2018robust} is randomly picked from the FACC1 \cite{Gabrilovich2013FACC1} data set, which provides annotations of mention-entity pairs for ClueWeb 2012 data\footnote{http://lemurproject.org/clueweb12}. 
  \item \textbf{WNED-WIKI} \cite{guo2018robust} is crawled from Wikipedia pages with its original anchor text annotations. Both WNED-CWEB and WNED-WIKI are automatically extracted by Guo and Barbosa \cite{guo2018robust} and are relatively large with a large number of documents.
\end{itemize}

\definecolor{Blue}{RGB}{80,190,255}
\begin{table}[t]
\small
  \centering
    \caption{Off-the-shelf EL tools.}  
  \label{tab:ELtools}
    \scalebox{0.9}{
  \begin{tabular}{l|p{6.3cm}}
    \toprule
    \textbf{EL tool} & \textbf{URL} \\ \midrule
    \textbf{StanfordCoreNLP}&   \url{https://stanfordnlp.github.io/CoreNLP/entitylink.html} \\ \hline
    \textbf{spaCy}&    \url{https://spacy.io/api/entitylinker} \\ \hline
    \textbf{TAGME}& \url{https://services.d4science.org/web/tagme/tagme-help}    \\ \hline
    \textbf{Wikipedia Miner}&  \url{http://community.nzdl.org/wikification/}   \\ \hline
    \textbf{DBpedia Spotlight}& \url{https://www.dbpedia-spotlight.org/}    \\ \hline
    \textbf{Babelfy}& \url{http://babelfy.org/}    \\  \hline
    \textbf{AGDISTIS}& \url{https://github.com/dice-group/AGDISTIS}    \\ \hline
    \textbf{WAT}&  \url{https://services.d4science.org/web/tagme/wat-api}   \\ \hline
    \textbf{FEL}&  \url{https://github.com/yahoo/FEL}   \\   \hline
    \textbf{REL}& \url{https://github.com/informagi/REL}    \\ 
 \bottomrule
  \end{tabular}}
  \vspace{-6mm}
\end{table}

\begin{table*}[t]
  \centering
  \caption{The performance of representative DL based EL methods on various data sets taken from both their original papers and the GERBIL platform. The best results are in \textbf{boldface} and the second-best results are \underline{underlined}. Due to the limited space, in the table header we show the abbreviations of the data sets which are defined in the first column of Table \ref{tb:datasets}. }  
  \label{tb:comparison}
  \scalebox{0.985}{
  \begin{tabular}{c|c|c|c|c|c|c|c|c|c|c}
    \hline
      \multirow{2}{*}{\textbf{Model}}  & \textbf{MSNBC} & \textbf{AQ} & \textbf{KBP} & \textbf{AIDA} & \textbf{ACE} & \textbf{KORE} & \textbf{RSS} & \textbf{Reuters} & \textbf{CW} & \textbf{WI} \\ \cline{2-11}
      ~ & $F1_{mic}$ & $F1_{mic}$ & $A_{mic}$ & $F1_{mic}$ & $F1_{mic}$ & $F1_{mic}$ & $F1_{mic}$ & $F1_{mic}$ & $F1_{mic}$ & $F1_{mic}$ \\ \hline
      \textbf{He et al.} (ACL 2013) \cite{he2013learning} & - & - & 81.0 & 85.6 & - & - & - & - & - & - \\ \hline
      \textbf{Sun et al.} (IJCAI 2015) \cite{sun2015modeling} & - & - & 83.9 & - & - & - & - & - & - & - \\ \hline
      \textbf{DSRM} (arXiv 2015) \cite{huang2015leveraging} & - & - & - & 86.6 & - & - & - & - & - & - \\ \hline
      \textbf{EDKate} (CoNLL 2016) \cite{fang2016entity} & 75.5 & 85.2 & 88.9 & - & 80.8 & - & - & - & - & - \\ \hline
      \textbf{Zwicklbauer et al.} (SIGIR 2016) \cite{zwicklbauer2016robust} & 91.1 & 84.2 & - & 78.4 & 90.7 & - & - & - & - & - \\ \hline
      \textbf{Francis-Landau et al.} (NAACL 2016) \cite{francis2016capturing} & - & 89.9 & - & 85.5 & - & - & - & - & - & - \\ \hline
      \textbf{Nguyen et al.} (COLING 2016) \cite{nguyen2016joint} & - & - & - & 87.2 & 89.7 & - & - & - & - & - \\ \hline
      \textbf{Globerson et al.} (ACL 2016) \cite{globerson2016collective} & - & - & 87.2 & 92.7 & - & - & - & - & - & - \\ \hline
      \textbf{Yamada et al.} (CoNLL 2016) \cite{yamada2016joint} & - & - & 85.5 & 93.1 & - & - & - & - & - & - \\ \hline
      \textbf{Gupta et al.} (EMNLP 2017) \cite{gupta2017entity} & - & - & - & 82.9 & 90.7 & - & - & - & - & - \\ \hline
      \textbf{Eshel et al.} (CoNLL 2017) \cite{eshel2017named} & - & - & - & 87.3 & - & - & - & - & - & - \\ \hline
      \textbf{Deep-ED} (EMNLP 2017) \cite{ganea2017deep} & 93.7 & 88.5 & - & 92.2 & 88.5 & - & - & - & 77.9 & 77.5 \\ \hline
      \textbf{NeuPL} (CIKM 2017) \cite{phan2017neupl} & 91.8 & - & - & - & \textbf{92.9} & \textbf{79.4} & \textbf{80.0} & \textbf{91.6} & - & - \\ \hline
      \textbf{NCEL} (COLING 2018) \cite{cao2018neural} & - & 87.0 & \underline{91.0} & 80.0 & 88.0 & - & - & - & - & \underline{86.0} \\ \hline
      \textbf{Kolitsas et al.} (CoNLL 2018) \cite{kolitsas2018end} & 86.4 & - & - & 83.1 & - & 60.8 & \underline{68.6} & 67.3 & - & - \\ \hline
      \textbf{SGTB-BiBSG} (NAACL 2018) \cite{yang2018collective} & 92.6 & 89.9 & - & 93.0 & 88.5 & - & - & - & \textbf{81.8} & 79.2 \\ \hline
      \textbf{MR-Deep-ED} (ACL 2018) \cite{le2018improving} & 93.9 & 88.3 & - & 93.1 & 89.9 & - & - & - & 77.5 & 78.0 \\ \hline
      \textbf{Sil et al.} (AAAI 2018) \cite{sil2017neural} & - & - & 87.4 & 94.0 & - & - &-  & - & - & - \\ \hline
      \textbf{DeepType} (AAAI 2018) \cite{raiman2018deeptype} & - & - & 90.9 & \underline{94.9} & - & - & - & - & - & - \\ \hline
      \textbf{Le and Titov} (ACL 2019) \cite{le2019distant} & - & - & - & 81.5 & - & - & - & - & - & - \\ \hline
      \textbf{Gillick et al.} (CoNLL 2019) \cite{gillick2019learning} & - & - & 87.0 & - & - & - & - & - & - & - \\ \hline
      \textbf{Le and Titov} (ACL 2019) \cite{le2019boosting} & 92.2 & 90.7 & - & 89.7 & 88.1 & - & - & - & 78.2 & 81.7 \\ \hline
      \textbf{RRWEL} (IJCAI 2019) \cite{xue2019neural} & 94.4 & \underline{91.9} & - & 92.4 & 90.6 & - & - & - & 79.7 & 85.5 \\ \hline
      \textbf{E-ELMo} (arXiv 2019) \cite{shahbazi2019entity} & 92.3 & 90.1 & 88.3 & 93.5 & 88.7 & - & - & - & 78.4 & 79.8 \\ \hline
      \textbf{RLEL} (WWW 2019) \cite{fang2019joint} & 92.8 & 87.5 & - & 94.3 & 91.2 & - & - & - & 78.5 & 82.8 \\ \hline
      \textbf{DCA-RL} (EMNLP 2019) \cite{yang2019learning} & 93.8 & 88.3 & - & 93.7 & 90.1 & - & - & - & 75.6 & 78.8 \\ \hline
      \textbf{DCA-SL} (EMNLP 2019) \cite{yang2019learning} & 94.6 & 87.4 & - & 94.6 & 89.4 & - & - & - & 73.5 & 78.2 \\ \hline
      \textbf{SeqGAT} (WWW 2020) \cite{fang2020high} & 80.0 & 88.0 & - & 83.0 & 89.0 & 68.0 & 68.0 & \underline{71.0} & - & - \\ \hline
      \textbf{REL} (SIGIR 2020) \cite{van2020rel} & 85.8 & - & - & 84.0 & - & 54.0 & 64.1 & 64.9 & - & - \\ \hline
      \textbf{ET4EL} (AAAI 2020) \cite{onoe2020fine} & - & - & - & 85.9 & - & - & - & - & - & - \\ \hline
      \textbf{FGS2EE} (ACL 2020) \cite{hou2020improving} & 94.2 & 88.5 & - & 92.6 & 90.7 & - & - & - & 77.4 & 77.8 \\ \hline
      \textbf{DGCN} (WWW 2020) \cite{wu2020dynamic} & 92.5 & 89.4 & - & 93.1 & 90.6 & - & - & - & \underline{81.2} & 77.6 \\ \hline
      \textbf{Chen et al.} (AAAI 2020) \cite{chen2020improving} & - & 89.8 & - & 93.4 & - & - & - & - & 77.9 & 80.1 \\ \hline
      \textbf{GNED} (KBS 2020) \cite{hu2020graph} & \underline{95.5} & 91.6 & - & 92.4 & 90.1 & - & - & - & 77.5 & 78.5 \\ \hline
      \textbf{BLINK} (EMNLP 2020) \cite{wu2020scalable} & - & - & \textbf{94.5} & - & - & - & - & - & - & - \\ \hline
      \textbf{Yamada et al.} (arXiv 2020) \cite{yamada2020global} & \textbf{96.3} & \textbf{93.5} & - & \textbf{95.0} & \underline{91.9} & - & - & - & 78.9 & \textbf{89.1} \\ \hline
      \textbf{M3} (AAAI 2021) \cite{gu2021read} & - & - & - & - & - & \underline{74.3} & - & - & - & - \\ \hline
      \textbf{CHOLAN} (EACL 2021) \cite{ravi2021cholan} & 83.4 & 76.8 & - & 85.7 & 86.8 & - & - & - & - & - \\ \hline
  \end{tabular}}
  \vspace{-2mm}
\end{table*}

In Table \ref{tb:datasets}, it can be seen that each data set has its own characteristics. For example, KORE50 \cite{hoffart2012kore} and N3-RSS500 \cite{roder2014n3} emphasize entity linking over short documents, which are composed of tweets and RSS-feeds respectively, while other data sets are constructed using relatively long news documents. What's more, besides being used for testing, some data sets, particularly larger ones like AIDA-CoNLL \cite{hoffart2011robust} and TAC-KBP2010 \cite{ji2011knowledge} could be also used for training, while some data sets such as MSNBC \cite{cucerzan2007large} and ACE2004 \cite{ratinov2011local} containing a few documents are generally not used for training. Thus, researchers could select appropriate data sets based on the different characteristics of their own EL systems when training or testing.

Lots of off-the-shelf entity linking tools are publicly available online. Table \ref{tab:ELtools} summarizes popular EL tools and their URLs.

\vspace{-2.5mm}
\subsection{Evaluation Metrics}
\label{metrics}

Evaluation metrics are used to evaluate the performance of EL systems on data sets. GERBIL \cite{usbeck2015gerbil} is a benchmark entity annotation platform that provides a unified comparison among different EL systems across various data sets and metrics. Currently, GERBIL offers six metrics and subdivides them into two groups, namely the macro- and the micro- groups of precision, recall, and F1-measure. The macro- metric is the average of the corresponding metric over each document in the data set $\mathcal{S}$, while the micro- metric takes into account all annotations together thus giving more importance to documents with more entity mentions \cite{cornolti2013framework}. The majority of DL based EL methods select micro- metrics for evaluation, whereas some EL works \cite{onoe2020fine,chen2020improving,he2013learning,cao2017bridge,huang2015leveraging,cao2018neural,yamada2016joint} utilize macro- metrics as well. Let $E_\mathcal{S}$, $E_D$ be the gold mapping entity annotations associated with all the entity mentions $M_{\mathcal{S}}$ in the data set $\mathcal{S}$, and a set of entity mentions $M_D$ in the document $D$ respectively. Let $G_\mathcal{S}$, $G_D$ be the output entity annotations of some EL system associated with all the entity mentions $M_{\mathcal{S}}$ in the data set $\mathcal{S}$, and a set of entity mentions $M_D$ in the document $D$ respectively. We give the definitions of the evaluation metrics, i.e., precision, recall, and F1-measure of both macro- and micro- groups as follows:  

\vspace{-1.5mm}
\begin{equation}
  \begin{split}
    &P_{mac} =  (\textstyle \sum_{D \in \mathcal{S}}^{}\frac{|E_D\cap G_D|}{|G_D|}) / |\mathcal{S}| \\
    &R_{mac} = (\textstyle \sum_{D \in \mathcal{S}}^{}\frac{|E_D\cap G_D|}{|E_D|}) / |\mathcal{S}| \\
    &F1_{mac} = 2 \cdot P_{mac} \cdot R_{mac} /(P_{mac} + R_{mac})\\
    &P_{mic} = |E_\mathcal{S}\cap G_\mathcal{S}|/|G_\mathcal{S}|\\
    &R_{mic} = |E_\mathcal{S}\cap G_\mathcal{S}|/|E_\mathcal{S}|\\
    &F1_{mic} = 2 \cdot P_{mic} \cdot R_{mic} /(P_{mic} + R_{mic})\\
  \end{split}
\end{equation}

\noindent The precision is computed as the fraction of correctly linked entity mentions that are generated by the EL system, and the recall is computed as the fraction of correctly linked entity mentions that should be correctly linked. To take into consideration both of them, the F1-measure puts them together, and it is defined as the harmonic mean of the precision and recall. In addition, the accuracy refers to the ratio of the number of entity mentions that are correctly linked to the total number of entity mentions in the data set, and it is defined as follows:

\begin{equation}
  \begin{split}
    &A_{mac} =  (\textstyle \sum_{D \in \mathcal{S}}^{}\frac{|E_D\cap G_D|}{|M_D|} )/{|\mathcal{S}|} \\
    &A_{mic} = |E_\mathcal{S}\cap G_\mathcal{S}| / |M_{\mathcal{S}}|  \\
  \end{split}
  \vspace{-2.5mm}
\end{equation}

\vspace{-2.5mm}
\subsection{Performance Analysis}
\label{comparison}

Table \ref{tb:comparison} presents the performance of representative DL based EL methods on data sets introduced in Section \ref{sec:data sets}. Most recent EL works used the GERBIL \cite{usbeck2015gerbil} platform to report their performance. To ensure correctness, we collect the experimental results from both their original papers and the GERBIL platform. The micro-F1 metric is commonly used by DL based EL methods to report the performance. Therefore, we show micro-F1 scores for all data sets except TAC-KBP2010 \cite{ji2011knowledge} since the micro-accuracy is regarded as the official evaluation metric in the TAC-KBP track.

In Table \ref{tb:comparison}, we can see that DL based EL methods have achieved the state-of-the-art performance on all data sets, which demonstrates the effectiveness of DL. Specifically, Yamada et al. \cite{yamada2020global} achieves the best results on four data sets (i.e., MSNBC \cite{cucerzan2007large}, AQUAINT \cite{milne2008learning}, AIDA-CoNLL \cite{hoffart2011robust}, and WNED-WIKI \cite{guo2018robust}), Wu et al. \cite{wu2020scalable} performs best on TAC-KBP2010 \cite{ji2011knowledge}, NeuPL \cite{phan2017neupl} owns the leadership on four data sets (i.e., ACE2004 \cite{ratinov2011local}, KORE50 \cite{hoffart2012kore}, N3-RSS500 \cite{roder2014n3}, and N3-Reuters128 \cite{roder2014n3}), and SGTB-BiBSG \cite{yang2018collective} obtains the best result on WNED-CWEB \cite{guo2018robust}.

It can be found from Table \ref{tb:comparison} that Transformers-based EL systems (e.g., Yamada et al. \cite{yamada2020global}, Wu et al. \cite{wu2020scalable}, and Chen et al. \cite{chen2020improving}) achieve advanced performance on many data sets as they are pre-trained on huge corpora and can generate more sophisticated long-distance feature representations for entity linking. The good performance of models leveraging type information (e.g., DeepType \cite{raiman2018deeptype}, FGS2EE \cite{hou2020improving}, and Chen et al. \cite{chen2020improving}) demonstrates the effectiveness of type information and points out a promising direction for entity linking. Specifically, DeepType \cite{raiman2018deeptype} that only leverages the prior popularity feature and the type similarity feature for linking achieves the second-best result on AIDA-CoNLL, which is amazing and enlightening.

It is also noted that no perfect EL system can achieve the best results on all data sets due to the different characteristics of various data sets, such as the document genre, the document length, and the number of entity mentions per document. That is, the best EL method on one data set may perform poorly on other data sets. For example, Transformers-based EL method Yamada et al. \cite{yamada2020global} obtains four best results and one second-best result over five data sets, but performs not well on WNED-CWEB since documents in this data set are significantly longer than documents in other data sets. There are approximately $1700$ words per document in the WNED-CWEB data set, which is much longer than the maximum word length that BERT can deal with (i.e., $512$ words). Nevertheless, SGTB-BiBSG \cite{yang2018collective} performs excellent on WNED-CWEB because this model designs various hand-crafted features capturing document-level contextual information. What's more, NeuPL \cite{phan2017neupl} performs best on data sets containing short documents such as KORE50 and N3-RSS500 since the pair-linking algorithm proposed in the NeuPL model iteratively identifies and resolves pairs of entity mentions without the requirement of much global information. Some EL methods (e.g., DCA-SL \cite{yang2019learning}, Deep-ED \cite{ganea2017deep}, and DGCN \cite{wu2020dynamic}) mainly based on the topical coherence feature perform well on data sets such as MSNBC and AIDA-CoNLL which have tens of entity mentions per document, because they can capture the global interdependence between candidate entities of entity mentions in a document well.

Overall, the entity linking task is highly data and domain dependent and it is unlikely that a technique dominates all others across all types of data sets. For a given data set for entity linking, we should leverage suitable embeddings, features, and algorithms to obtain advanced results based on the characteristics of the data.

\vspace{-2.5mm}
\section{Future Directions}
\label{future}

Based on the above review and analysis, we believe that there is still much space for further enhancement in this field. In this section, we discuss the remaining limitations of existing EL methods and list some directions for further exploration in EL research.

\paratitle{Multi-source heterogeneous text data.} 
In the era of big data, text data have multi-source heterogeneous characteristics. Text may come from diverse data sources in various structures. For example, news documents from news websites are relatively long and formal. Web tables shown in web pages are structured. Queries from search engine logs are often short and noncontiguous. Reviews from e-commerce websites are usually colloquial and noisy. Entities may appear in those multi-source heterogeneous text data which contain abundant knowledge about them. Bridging the multi-source heterogeneous text data with KBs is beneficial for the understanding of the text data and the enrichment of KBs. Present EL studies mainly focus on linking entity mentions in common text data (e.g., news documents \cite{ganea2017deep,le2018improving,yamada2016joint,yamada2020global}, tweets \cite{guo2013link,shen2021toward,hua2015microblog}, and web tables \cite{zhang2020novel,ritze2016profiling,bhagavatula2015tabel}). As different types of text data have various characteristics, existing EL methods may not be applicable or be difficult to achieve satisfactory linking performance when dealing with other types of text data (e.g., search queries, reviews, community question answering (CQA) text, and open information extraction (OIE) triples). Therefore, it is very meaningful and essential to develop EL techniques to link entities in these diverse types. Although some works have preliminarily addressed the entity linking task for search queries \cite{tan2017entity,blanco2015fast,cornolti2016piggyback}, CQA text \cite{wang2017named}, and OIE triples \cite{lin2020kbpearl,liu2021joint} respectively, we believe there are still many opportunities for substantial improvement.

\paratitle{Joint NER and EL.}
NER is the task to identify text spans that mention named entities, and to classify them into pre-defined categories, such as person, location, and organization \cite{li2020survey}. It serves as a preceded task for entity linking, which provides the boundaries of named entities in text. However, detecting the correct entity mentions is challenging, especially for informal short noisy text that often contains phrases with ambiguous meanings. Therefore, NER is often the performance bottleneck of EL since the performance of NER leads to an upper limit to the performance of EL. Some EL methods are designed to jointly perform NER and EL, which allows each subtask to benefit from another and alleviates error propagations that are unavoidable in pipeline settings. So far, Guo et al. \cite{guo2013link} utilized a structured support vector machines algorithm for tweet entity linking that jointly optimizes NER and EL as an end-to-end task. Kolitsas et al. \cite{kolitsas2018end} and Broscheit \cite{broscheit2019investigating} proposed neural end-to-end EL systems that jointly discover and link entities in the news documents. Li et al. \cite{li2020efficient} designed an end-to-end EL system used for downstream question answering systems. In summary, we consider it is worth exploring effective approaches for jointly performing NER and EL for real applications in the future.

\paratitle{More advanced language models.}
Neural language models have revolutionized the field of NLP due to their superior expressive power. As introduced earlier, many effective language models have been applied in the field of EL and achieved great success. Recently, there are many more advanced language models being developed and available. For instance, BERT \cite{devlin2019bert}, widely leveraged by existing EL works, has been exceeded by several variants and other transformer-based models, which made major changes to loss functions, model architecture, and pre-training objectives. Specifically, RoBERTa \cite{liu2019roberta} is more robust than BERT which is trained using much more training data and leveraging dynamic masking rather than static masking. SpanBERT \cite{joshi2020spanbert} extends BERT to better represent and predict text spans. XLNet \cite{yang2019xlnet} is a generalized order-aware autoregressive language model which makes use of a permutation operation to perform better than BERT on lots of NLP tasks. Moreover, GPT-3 \cite{brown2020language} is the largest pre-trained language model by far with hundreds of billions of parameters, which has shown great abilities in zero-shot, one-shot, and few-shot settings. In summary, we consider there is still much space for further enhancement of EL by leveraging these more advanced language models in modeling text semantics.
 

\paratitle{EL model robustness.}
The robustness of deep learning has received great attention recently. A DL model is considered to be robust if its output label is consistently accurate even if one or more of the input features or assumptions are drastically changed due to unforeseen circumstances. For entity linking, robustness refers to the ability to achieve consistent performance over a wide range of data sets with different properties, such as different domains, text structures, and knowledge bases \cite{zwicklbauer2016robust}. However, most existing DL based EL systems were designed and optimized for a specific domain (e.g., general domain or biomedical domain), for a specific text structure (e.g., news document or tweet), and for a specific knowledge base (e.g., Wikipedia or Freebase). This leads to very specialized models that lack robustness and are applicable for very specific tasks. Therefore, we strongly believe that robust DL based EL models that generalize well deserve much deeper exploration by the community.

\vspace{-2.5mm}
\section{Conclusion}
\label{conclusion}

Applying deep learning to entity linking has become a popular research topic today. In this survey, we give a comprehensive and detailed review of the existing DL based EL methods. We first propose a new taxonomy which categories DL techniques for ranking candidate entities based on three axes (i.e., embedding, feature, and algorithm). Second, we systematically review the representative DL based EL methods according to the taxonomy. Third, we present ten widely used real-world entity linking data sets and a quantitative performance analysis of DL based EL methods in tabular form. Finally, we discuss some limitations and highlight several future research directions. Through this paper, we hope to demonstrate the progress and problem in existing entity linking research and encourage more improvements in this area.

\vspace{-0.5mm}
  
\vspace{-2.5mm}
\ifCLASSOPTIONcompsoc
  \section*{Acknowledgments}
  This work was supported in part by National Natural Science Foundation of China under Grant No. U1936206 and 61772289, Natural Science Foundation of Tianjin under Grant No. 19JCQNJC00100, YESS by CAST under Grant No. 2019QNRC001, and CAAI-Huawei MindSpore Open Fund.
Jianyong Wang was supported in part by National Key Research and Development Program of China under Grant No. 2020YFA0804503, National Natural Science Foundation of China under Grant No. 61532010 and 61521002, and Beijing Academy of Artificial Intelligence (BAAI).
\else
  \section*{Acknowledgment}
\fi

\vspace{-2.5mm}

\bibliographystyle{IEEEtran}
\bibliography{IEEEabrv,mybib}{}

\begin{thebibliography}{100}
\providecommand{\url}[1]{#1}
\csname url@samestyle\endcsname
\providecommand{\newblock}{\relax}
\providecommand{\bibinfo}[2]{#2}
\providecommand{\BIBentrySTDinterwordspacing}{\spaceskip=0pt\relax}
\providecommand{\BIBentryALTinterwordstretchfactor}{4}
\providecommand{\BIBentryALTinterwordspacing}{\spaceskip=\fontdimen2\font plus
\BIBentryALTinterwordstretchfactor\fontdimen3\font minus
  \fontdimen4\font\relax}
\providecommand{\BIBforeignlanguage}[2]{{%
\expandafter\ifx\csname l@#1\endcsname\relax
\typeout{** WARNING: IEEEtran.bst: No hyphenation pattern has been}%
\typeout{** loaded for the language `#1'. Using the pattern for}%
\typeout{** the default language instead.}%
\else
\language=\csname l@#1\endcsname
\fi
#2}}
\providecommand{\BIBdecl}{\relax}
\BIBdecl

\bibitem{suchanek2007yago}
F.~M. Suchanek, G.~Kasneci, and G.~Weikum, ``Yago: a core of semantic
  knowledge,'' in \emph{WWW}, 2007, pp. 697--706.

\bibitem{auer2007dbpedia}
S.~Auer, C.~Bizer, G.~Kobilarov, J.~Lehmann, R.~Cyganiak, and Z.~G. Ives,
  ``Dbpedia: {A} nucleus for a web of open data,'' in \emph{ISWC}, 2007, pp.
  722--735.

\bibitem{bollacker2008freebase}
K.~Bollacker, C.~Evans, P.~Paritosh, T.~Sturge, and J.~Taylor, ``Freebase: a
  collaboratively created graph database for structuring human knowledge,'' in
  \emph{SIGMOD}, 2008, pp. 1247--1250.

\bibitem{wu2012probase}
W.~Wu, H.~Li, H.~Wang, and K.~Q. Zhu, ``Probase: A probabilistic taxonomy for
  text understanding,'' in \emph{SIGMOD}, 2012, pp. 481--492.

\bibitem{shen2014entity}
W.~Shen, J.~Wang, and J.~Han, ``Entity linking with a knowledge base: Issues,
  techniques, and solutions,'' \emph{TKDE}, vol.~27, no.~2, pp. 443--460, 2014.

\bibitem{zhang2016joint}
Y.~Zhang, S.~He, K.~Liu, and J.~Zhao, ``A joint model for question answering
  over multiple knowledge bases,'' in \emph{AAAI}, 2016, pp. 3094--3100.

\bibitem{lin2016neural}
Y.~Lin, S.~Shen, Z.~Liu, H.~Luan, and M.~Sun, ``Neural relation extraction with
  selective attention over instances,'' in \emph{ACL}, 2016, pp. 2124--2133.

\bibitem{ji2011knowledge}
H.~Ji and R.~Grishman, ``Knowledge base population: Successful approaches and
  challenges,'' in \emph{ACL}, 2011, pp. 1148--1158.

\bibitem{michelson2010discovering}
M.~Michelson and S.~A. Macskassy, ``Discovering users' topics of interest on
  twitter: a first look,'' in \emph{Proceedings of the fourth workshop on
  Analytics for noisy unstructured text data}, 2010, pp. 73--80.

\bibitem{cucerzan2007large}
S.~Cucerzan, ``Large-scale named entity disambiguation based on wikipedia
  data,'' in \emph{EMNLP-CoNLL}, 2007, pp. 708--716.

\bibitem{varma2009iiit}
V.~Varma, P.~Pingali, R.~Katragadda, S.~Krishna, S.~Ganesh, K.~Sarvabhotla,
  H.~Garapati, H.~Gopisetty, V.~B. Reddy, K.~Reddy \emph{et~al.}, ``Iiit
  hyderabad at tac 2009.'' in \emph{Text Analysis Conference 2009 Workshop},
  2009.

\bibitem{chen2011collaborative}
Z.~Chen and H.~Ji, ``Collaborative ranking: A case study on entity linking,''
  in \emph{EMNLP}, 2011, pp. 771--781.

\bibitem{ratinov2011local}
L.~Ratinov, D.~Roth, D.~Downey, and M.~Anderson, ``Local and global algorithms
  for disambiguation to wikipedia,'' in \emph{ACL}, 2011, pp. 1375--1384.

\bibitem{shen2017shine+}
W.~Shen, J.~Han, J.~Wang, X.~Yuan, and Z.~Yang, ``Shine+: A general framework
  for domain-specific entity linking with heterogeneous information networks,''
  \emph{TKDE}, vol.~30, no.~2, pp. 353--366, 2017.

\bibitem{hoffart2011robust}
J.~Hoffart, M.~A. Yosef, I.~Bordino, H.~F{\"u}rstenau, M.~Pinkal, M.~Spaniol,
  B.~Taneva, S.~Thater, and G.~Weikum, ``Robust disambiguation of named
  entities in text,'' in \emph{EMNLP}, 2011, pp. 782--792.

\bibitem{sevgili2020neural}
O.~Sevgili, A.~Shelmanov, M.~Arkhipov, A.~Panchenko, and C.~Biemann, ``Neural
  entity linking: A survey of models based on deep learning,'' \emph{arXiv
  preprint arXiv:2006.00575}, 2020.

\bibitem{mikolov2013efficient}
T.~Mikolov, K.~Chen, G.~Corrado, and J.~Dean, ``Efficient estimation of word
  representations in vector space,'' in \emph{ICLR}, 2013.

\bibitem{mikolov2013distributed}
T.~Mikolov, I.~Sutskever, K.~Chen, G.~S. Corrado, and J.~Dean, ``Distributed
  representations of words and phrases and their compositionality,'' in
  \emph{NIPS}, 2013, pp. 3111--3119.

\bibitem{pennington2014glove}
J.~Pennington, R.~Socher, and C.~D. Manning, ``Glove: Global vectors for word
  representation,'' in \emph{EMNLP}, 2014, pp. 1532--1543.

\bibitem{hochreiter1997long}
S.~Hochreiter and J.~Schmidhuber, ``Long short-term memory,'' \emph{Neural
  Computation}, vol.~9, no.~8, pp. 1735--1780, 1997.

\bibitem{cho2014learning}
K.~Cho, B.~van Merri{\"e}nboer, C.~Gulcehre, D.~Bahdanau, F.~Bougares,
  H.~Schwenk, and Y.~Bengio, ``Learning phrase representations using rnn
  encoder--decoder for statistical machine translation,'' in \emph{EMNLP},
  2014, pp. 1724--1734.

\bibitem{lecun1998gradient}
Y.~LeCun, L.~Bottou, Y.~Bengio, and P.~Haffner, ``Gradient-based learning
  applied to document recognition,'' \emph{Proceedings of the IEEE}, vol.~86,
  no.~11, pp. 2278--2324, 1998.

\bibitem{bahdanau2015neural}
D.~Bahdanau, K.~Cho, and Y.~Bengio, ``Neural machine translation by jointly
  learning to align and translate,'' in \emph{ICLR}, 2015.

\bibitem{vaswani2017attention}
A.~Vaswani, N.~Shazeer, N.~Parmar, J.~Uszkoreit, L.~Jones, A.~N. Gomez,
  {\L}.~Kaiser, and I.~Polosukhin, ``Attention is all you need,'' in
  \emph{NIPS}, 2017, pp. 5998--6008.

\bibitem{krizhevsky2012imagenet}
A.~Krizhevsky, I.~Sutskever, and G.~E. Hinton, ``Imagenet classification with
  deep convolutional neural networks,'' in \emph{NIPS}, 2012, pp. 1097--1105.

\bibitem{mudgal2018deep}
S.~Mudgal, H.~Li, T.~Rekatsinas, A.~Doan, Y.~Park, G.~Krishnan, R.~Deep,
  E.~Arcaute, and V.~Raghavendra, ``Deep learning for entity matching: A design
  space exploration,'' in \emph{SIGMOD}, 2018, pp. 19--34.

\bibitem{glorot2011deep}
X.~Glorot, A.~Bordes, and Y.~Bengio, ``Deep sparse rectifier neural networks,''
  in \emph{AISTATS}, 2011, pp. 315--323.

\bibitem{bengio2013deep}
Y.~Bengio, ``Deep learning of representations: Looking forward,'' in
  \emph{SLPS}, 2013, pp. 1--37.

\bibitem{wang2018deep}
M.~Wang and W.~Deng, ``Deep visual domain adaptation: A survey,''
  \emph{Neurocomputing}, vol. 312, pp. 135--153, 2018.

\bibitem{collobert2011natural}
R.~Collobert, J.~Weston, L.~Bottou, M.~Karlen, K.~Kavukcuoglu, and P.~Kuksa,
  ``Natural language processing (almost) from scratch,'' \emph{JMLR}, vol.~12,
  no. ARTICLE, pp. 2493--2537, 2011.

\bibitem{wu2020dynamic}
J.~Wu, R.~Zhang, Y.~Mao, H.~Guo, M.~Soflaei, and J.~Huai, ``Dynamic graph
  convolutional networks for entity linking,'' in \emph{WWW}, 2020, pp.
  1149--1159.

\bibitem{he2013learning}
Z.~He, S.~Liu, M.~Li, M.~Zhou, L.~Zhang, and H.~Wang, ``Learning entity
  representation for entity disambiguation,'' in \emph{ACL}, 2013, pp. 30--34.

\bibitem{sun2015modeling}
Y.~Sun, L.~Lin, D.~Tang, N.~Yang, Z.~Ji, and X.~Wang, ``Modeling mention,
  context and entity with neural networks for entity disambiguation.'' in
  \emph{IJCAI}, 2015, pp. 1333--1339.

\bibitem{huang2015leveraging}
H.~Huang, L.~Heck, and H.~Ji, ``Leveraging deep neural networks and knowledge
  graphs for entity disambiguation,'' \emph{arXiv preprint arXiv:1504.07678},
  2015.

\bibitem{globerson2016collective}
A.~Globerson, N.~Lazic, S.~Chakrabarti, A.~Subramanya, M.~Ringgaard, and
  F.~Pereira, ``Collective entity resolution with multi-focal attention,'' in
  \emph{ACL}, 2016, pp. 621--631.

\bibitem{zwicklbauer2016robust}
S.~Zwicklbauer, C.~Seifert, and M.~Granitzer, ``Robust and collective entity
  disambiguation through semantic embeddings,'' in \emph{SIGIR}, 2016, pp.
  425--434.

\bibitem{fang2016entity}
W.~Fang, J.~Zhang, D.~Wang, Z.~Chen, and M.~Li, ``Entity disambiguation by
  knowledge and text jointly embedding,'' in \emph{CoNLL}, 2016, pp. 260--269.

\bibitem{yamada2016joint}
I.~Yamada, H.~Shindo, H.~Takeda, and Y.~Takefuji, ``Joint learning of the
  embedding of words and entities for named entity disambiguation,'' in
  \emph{CoNLL}, 2016, pp. 250--259.

\bibitem{francis2016capturing}
M.~Francis-Landau, G.~Durrett, and D.~Klein, ``Capturing semantic similarity
  for entity linking with convolutional neural networks,'' in \emph{NAACL},
  2016, pp. 1256--1261.

\bibitem{nguyen2016joint}
T.~H. Nguyen, N.~R. Fauceglia, M.~R. Muro, O.~Hassanzadeh, A.~Gliozzo, and
  M.~Sadoghi, ``Joint learning of local and global features for entity linking
  via neural networks,'' in \emph{COLING}, 2016, pp. 2310--2320.

\bibitem{cao2017bridge}
Y.~Cao, L.~Huang, H.~Ji, X.~Chen, and J.~Li, ``Bridge text and knowledge by
  learning multi-prototype entity mention embedding,'' in \emph{ACL}, 2017, pp.
  1623--1633.

\bibitem{gupta2017entity}
N.~Gupta, S.~Singh, and D.~Roth, ``Entity linking via joint encoding of types,
  descriptions, and context,'' in \emph{EMNLP}, 2017, pp. 2681--2690.

\bibitem{ganea2017deep}
O.-E. Ganea and T.~Hofmann, ``Deep joint entity disambiguation with local
  neural attention,'' in \emph{EMNLP}, 2017, pp. 2619--2629.

\bibitem{phan2017neupl}
M.~C. Phan, A.~Sun, Y.~Tay, J.~Han, and C.~Li, ``Neupl: Attention-based
  semantic matching and pair-linking for entity disambiguation,'' in
  \emph{CIKM}, 2017, pp. 1667--1676.

\bibitem{eshel2017named}
Y.~Eshel, N.~Cohen, K.~Radinsky, S.~Markovitch, I.~Yamada, and O.~Levy, ``Named
  entity disambiguation for noisy text,'' in \emph{CoNLL}, 2017, pp. 58--68.

\bibitem{le2018improving}
P.~Le and I.~Titov, ``Improving entity linking by modeling latent relations
  between mentions,'' in \emph{ACL}, 2018, pp. 1595--1604.

\bibitem{moon2018multimodal}
S.~Moon, L.~Neves, and V.~Carvalho, ``Multimodal named entity disambiguation
  for noisy social media posts,'' in \emph{ACL}, 2018, pp. 2000--2008.

\bibitem{sil2017neural}
A.~Sil, G.~Kundu, R.~Florian, and W.~Hamza, ``Neural cross-lingual entity
  linking,'' in \emph{AAAI}, 2017, pp. 5464--5472.

\bibitem{raiman2018deeptype}
J.~Raiman and O.~Raiman, ``Deeptype: Multilingual entity linking by neural type
  system evolution,'' in \emph{AAAI}, 2018, pp. 5406--5413.

\bibitem{mueller2018effective}
D.~Mueller and G.~Durrett, ``Effective use of context in noisy entity
  linking,'' in \emph{EMNLP}, 2018, pp. 1024--1029.

\bibitem{kolitsas2018end}
N.~Kolitsas, O.-E. Ganea, and T.~Hofmann, ``End-to-end neural entity linking,''
  in \emph{CoNLL}, 2018, pp. 519--529.

\bibitem{yang2018collective}
Y.~Yang, O.~{\.I}rsoy, and K.~S. Rahman, ``Collective entity disambiguation
  with structured gradient tree boosting,'' in \emph{NAACL}, 2018, pp.
  777--786.

\bibitem{cao2018neural}
Y.~Cao, L.~Hou, J.~Li, and Z.~Liu, ``Neural collective entity linking,'' in
  \emph{COLING}, 2018, pp. 675--686.

\bibitem{le2019distant}
P.~Le and I.~Titov, ``Distant learning for entity linking with automatic noise
  detection,'' in \emph{ACL}, 2019, pp. 4081--4090.

\bibitem{le2019boosting}
P.~Le and I.~Titov, ``Boosting entity linking performance by leveraging
  unlabeled documents,'' in \emph{ACL}, 2019, pp. 1935--1945.

\bibitem{logeswaran2019zero}
L.~Logeswaran, M.-W. Chang, K.~Lee, K.~Toutanova, J.~Devlin, and H.~Lee,
  ``Zero-shot entity linking by reading entity descriptions,'' in \emph{ACL},
  2019, pp. 3449--3460.

\bibitem{sevgili2019improving}
{\"O}.~Sevgili, A.~Panchenko, and C.~Biemann, ``Improving neural entity
  disambiguation with graph embeddings,'' in \emph{ACL}, 2019, pp. 315--322.

\bibitem{xue2019neural}
M.~Xue, W.~Cai, J.~Su, L.~Song, Y.~Ge, Y.~Liu, and B.~Wang, ``Neural collective
  entity linking based on recurrent random walk network learning,'' in
  \emph{IJCAI}, 2019, pp. 5327--5333.

\bibitem{fang2019joint}
Z.~Fang, Y.~Cao, Q.~Li, D.~Zhang, Z.~Zhang, and Y.~Liu, ``Joint entity linking
  with deep reinforcement learning,'' in \emph{WWW}, 2019, pp. 438--447.

\bibitem{yang2019learning}
X.~Yang, X.~Gu, S.~Lin, S.~Tang, Y.~Zhuang, F.~Wu, Z.~Chen, G.~Hu, and X.~Ren,
  ``Learning dynamic context augmentation for global entity linking,'' in
  \emph{EMNLP-IJCNLP}, 2019, pp. 271--281.

\bibitem{gillick2019learning}
D.~Gillick, S.~Kulkarni, L.~Lansing, A.~Presta, J.~Baldridge, E.~Ie, and
  D.~Garcia-Olano, ``Learning dense representations for entity retrieval,'' in
  \emph{CoNLL}, 2019, pp. 528--537.

\bibitem{shahbazi2019entity}
H.~Shahbazi, X.~Z. Fern, R.~Ghaeini, R.~Obeidat, and P.~Tadepalli,
  ``Entity-aware elmo: Learning contextual entity representation for entity
  disambiguation,'' \emph{arXiv preprint arXiv:1908.05762}, 2019.

\bibitem{hou2020improving}
F.~Hou, R.~Wang, J.~He, and Y.~Zhou, ``Improving entity linking through
  semantic reinforced entity embeddings,'' in \emph{ACL}, 2020, pp. 6843--6848.

\bibitem{onoe2020fine}
Y.~Onoe and G.~Durrett, ``Fine-grained entity typing for domain independent
  entity linking.'' in \emph{AAAI}, 2020, pp. 8576--8583.

\bibitem{chen2020improving}
S.~Chen, J.~Wang, F.~Jiang, and C.-Y. Lin, ``Improving entity linking by
  modeling latent entity type information,'' in \emph{AAAI}, 2020, pp.
  7529--7537.

\bibitem{van2020rel}
J.~M. van Hulst, F.~Hasibi, K.~Dercksen, K.~Balog, and A.~P. de~Vries, ``Rel:
  An entity linker standing on the shoulders of giants,'' in \emph{SIGIR},
  2020, pp. 2197--2200.

\bibitem{fang2020high}
Z.~Fang, Y.~Cao, R.~Li, Z.~Zhang, Y.~Liu, and S.~Wang, ``High quality candidate
  generation and sequential graph attention network for entity linking,'' in
  \emph{WWW}, 2020, pp. 640--650.

\bibitem{wu2020scalable}
L.~Wu, F.~Petroni, M.~Josifoski, S.~Riedel, and L.~Zettlemoyer, ``Scalable
  zero-shot entity linking with dense entity retrieval,'' in \emph{EMNLP},
  2020, pp. 6397--6407.

\bibitem{li2020efficient}
B.~Z. Li, S.~Min, S.~Iyer, Y.~Mehdad, and W.-t. Yih, ``Efficient one-pass
  end-to-end entity linking for questions,'' in \emph{EMNLP}, 2020, pp.
  6433--6441.

\bibitem{hu2020graph}
L.~Hu, J.~Ding, C.~Shi, C.~Shao, and S.~Li, ``Graph neural entity
  disambiguation,'' \emph{KBS}, vol. 195, p. 105620, 2020.

\bibitem{adjali2020ecir}
O.~Adjali, r.~Besancon, o.~Ferret, H.~{Le Borgne}, and B.~Grau, ``Multimodal
  entity linking for tweets,'' in \emph{ECIR}, 2020.

\bibitem{yamada2020global}
I.~Yamada, K.~Washio, H.~Shindo, and Y.~Matsumoto, ``Global entity
  disambiguation with pretrained contextualized embeddings of words and
  entities,'' \emph{arXiv preprint arXiv:1909.00426}, 2020.

\bibitem{gu2021read}
Y.~Gu, X.~Qu, Z.~Wang, B.~Huai, N.~J. Yuan, and X.~Gui, ``Read, retrospect,
  select: An mrc framework to short text entity linking,'' in \emph{AAAI},
  2021, pp. 12\,920--12\,928.

\bibitem{tang2021bidirectional}
H.~Tang, X.~Sun, B.~Jin, and F.~Zhang, ``A bidirectional multi-paragraph
  reading model for zero-shot entity linking,'' in \emph{AAAI}, 2021, pp.
  13\,889--13\,897.

\bibitem{chen2021lightweight}
L.~Chen, G.~Varoquaux, and F.~M. Suchanek, ``A lightweight neural model for
  biomedical entity linking,'' in \emph{AAAI}, 2021, pp. 12\,657--12\,665.

\bibitem{ravi2021cholan}
M.~P.~K. Ravi, K.~Singh, I.~O. Mulang, S.~Shekarpour, J.~Hoffart, and
  J.~Lehmann, ``Cholan: A modular approach for neural entity linking on
  wikipedia and wikidata,'' in \emph{EACL}, 2021, pp. 504--514.

\bibitem{zhang2021attention}
L.~Zhang, Z.~Li, and Q.~Yang, ``Attention-based multimodal entity linking with
  high-quality images,'' in \emph{DASFAA}, 2021, pp. 533--548.

\bibitem{phan2018pair}
M.~C. Phan, A.~Sun, Y.~Tay, J.~Han, and C.~Li, ``Pair-linking for collective
  entity disambiguation: Two could be better than all,'' \emph{TKDE}, vol.~31,
  no.~7, pp. 1383--1396, 2018.

\bibitem{peters2018deep}
M.~E. Peters, M.~Neumann, M.~Iyyer, M.~Gardner, C.~Clark, K.~Lee, and
  L.~Zettlemoyer, ``Deep contextualized word representations,'' in
  \emph{NAACL}, 2018, pp. 2227--2237.

\bibitem{lample2016neural}
G.~Lample, M.~Ballesteros, S.~Subramanian, K.~Kawakami, and C.~Dyer, ``Neural
  architectures for named entity recognition,'' in \emph{NAACL}, 2016, pp.
  260--270.

\bibitem{pagliardini2018unsupervised}
M.~Pagliardini, P.~Gupta, and M.~Jaggi, ``Unsupervised learning of sentence
  embeddings using compositional n-gram features,'' in \emph{NAACL}, 2018, pp.
  528--540.

\bibitem{kim2016character}
Y.~Kim, Y.~Jernite, D.~Sontag, and A.~M. Rush, ``Character-aware neural
  language models,'' in \emph{AAAI}, 2016, pp. 2741--2749.

\bibitem{devlin2019bert}
J.~Devlin, M.-W. Chang, K.~Lee, and K.~Toutanova, ``Bert: Pre-training of deep
  bidirectional transformers for language understanding,'' in \emph{NAACL},
  2019, pp. 4171--4186.

\bibitem{charton2014improving}
E.~Charton, M.-J. Meurs, L.~Jean-Louis, and M.~Gagnon, ``Improving entity
  linking using surface form refinement.'' in \emph{LREC}, 2014, pp.
  4609--4615.

\bibitem{vincent2008extracting}
P.~Vincent, H.~Larochelle, Y.~Bengio, and P.-A. Manzagol, ``Extracting and
  composing robust features with denoising autoencoders,'' in \emph{ICML},
  2008, pp. 1096--1103.

\bibitem{le2014distributed}
Q.~Le and T.~Mikolov, ``Distributed representations of sentences and
  documents,'' in \emph{ICML}, 2014, pp. 1188--1196.

\bibitem{kipf2016semi}
T.~N. Kipf and M.~Welling, ``Semi-supervised classification with graph
  convolutional networks,'' in \emph{ICLR}, 2016.

\bibitem{milne2008learning}
D.~Milne and I.~H. Witten, ``Learning to link with wikipedia,'' in \emph{CIKM},
  2008, pp. 509--518.

\bibitem{perozzi2014deepwalk}
B.~Perozzi, R.~Al-Rfou, and S.~Skiena, ``Deepwalk: Online learning of social
  representations,'' in \emph{SIGKDD}, 2014, pp. 701--710.

\bibitem{bordes2013translating}
A.~Bordes, N.~Usunier, A.~Garcia-Duran, J.~Weston, and O.~Yakhnenko,
  ``Translating embeddings for modeling multi-relational data,'' in
  \emph{NIPS}, 2013, pp. 1--9.

\bibitem{socher2013recursive}
R.~Socher, A.~Perelygin, J.~Wu, J.~Chuang, C.~D. Manning, A.~Y. Ng, and
  C.~Potts, ``Recursive deep models for semantic compositionality over a
  sentiment treebank,'' in \emph{EMNLP}, 2013, pp. 1631--1642.

\bibitem{levy2014dependency}
O.~Levy and Y.~Goldberg, ``Dependency-based word embeddings,'' in \emph{ACL},
  2014, pp. 302--308.

\bibitem{xu2018neural}
P.~Xu and D.~Barbosa, ``Neural fine-grained entity type classification with
  hierarchy-aware loss,'' in \emph{NAACL}, 2018, pp. 16--25.

\bibitem{guo2013link}
S.~Guo, M.-W. Chang, and E.~Kiciman, ``To link or not to link? a study on
  end-to-end tweet entity linking,'' in \emph{NAACL}, 2013, pp. 1020--1030.

\bibitem{zhang2010nus}
W.~Zhang, C.~L. Tan, Y.~C. Sim, and J.~Su, ``Nus-i2r: Learning a combined
  system for entity linking.'' in \emph{Text Analysis Conference 2010
  Workshop}, 2010.

\bibitem{vstajner2009entity}
T.~{\v{S}}tajner and D.~Mladeni{\'c}, ``Entity resolution in texts using
  statistical learning and ontologies,'' in \emph{ASWC}, 2009, pp. 91--104.

\bibitem{hoffart2012kore}
J.~Hoffart, S.~Seufert, D.~B. Nguyen, M.~Theobald, and G.~Weikum, ``Kore:
  keyphrase overlap relatedness for entity disambiguation,'' in \emph{CIKM},
  2012, pp. 545--554.

\bibitem{minaee2020deep}
S.~Minaee, N.~Kalchbrenner, E.~Cambria, N.~Nikzad, M.~Chenaghlu, and J.~Gao,
  ``Deep learning based text classification: A comprehensive review,''
  \emph{arXiv preprint arXiv:2004.03705}, 2020.

\bibitem{shen2012linden}
W.~Shen, J.~Wang, P.~Luo, and M.~Wang, ``Linden: linking named entities with
  knowledge base via semantic knowledge,'' in \emph{WWW}, 2012, pp. 449--458.

\bibitem{shen2012liege}
W.~Shen, J.~Wang, P.~Luo, and M.~Wang, ``Liege: link entities in web lists with
  knowledge base,'' in \emph{SIGKDD}, 2012, pp. 1424--1432.

\bibitem{pilz2011names}
A.~Pilz and G.~Paa{\ss}, ``From names to entities using thematic context
  distance,'' in \emph{CIKM}, 2011, pp. 857--866.

\bibitem{monahan2011cross}
S.~Monahan, J.~Lehmann, T.~Nyberg, J.~Plymale, and A.~Jung, ``Cross-lingual
  cross-document coreference with entity linking.'' in \emph{Text Analysis
  Conference 2011 Workshop}, 2011.

\bibitem{roder2014n3}
M.~R{\"o}der, R.~Usbeck, S.~Hellmann, D.~Gerber, and A.~Both, ``N$^3$-a
  collection of datasets for named entity recognition and disambiguation in the
  nlp interchange format.'' in \emph{LREC}, 2014, pp. 3529--3533.

\bibitem{guo2018robust}
Z.~Guo and D.~Barbosa, ``Robust named entity disambiguation with random
  walks,'' \emph{Semantic Web}, vol.~9, no.~4, pp. 459--479, 2018.

\bibitem{kaelbling1996reinforcement}
L.~P. Kaelbling, M.~L. Littman, and A.~W. Moore, ``Reinforcement learning: A
  survey,'' \emph{JAIR}, vol.~4, pp. 237--285, 1996.

\bibitem{doddington2004automatic}
G.~R. Doddington, A.~Mitchell, M.~Przybocki, L.~Ramshaw, S.~Strassel, and
  R.~Weischedel, ``The automatic content extraction (ace) program--tasks, data,
  and evaluation,'' in \emph{LREC}, 2004.

\bibitem{Gabrilovich2013FACC1}
E.~Gabrilovich, M.~Ringgaard, and A.~Subramanya, ``Facc1: Freebase annotation
  of clueweb corpora, version 1 (release date 2013-06-26, format version 1,
  correction level 0),'' 2013.

\bibitem{usbeck2015gerbil}
R.~Usbeck, M.~R{\"o}der, A.-C. Ngonga~Ngomo, C.~Baron, A.~Both, M.~Br{\"u}mmer,
  D.~Ceccarelli, M.~Cornolti, D.~Cherix, B.~Eickmann \emph{et~al.}, ``Gerbil:
  general entity annotator benchmarking framework,'' in \emph{WWW}, 2015, pp.
  1133--1143.

\bibitem{cornolti2013framework}
M.~Cornolti, P.~Ferragina, and M.~Ciaramita, ``A framework for benchmarking
  entity-annotation systems,'' in \emph{WWW}, 2013, pp. 249--260.

\bibitem{shen2021toward}
W.~Shen, Y.~Yin, Y.~Yang, J.~Han, J.~Wang, and X.~Yuan, ``Toward tweet entity
  linking with heterogeneous information networks,'' \emph{TKDE}, 2021.

\bibitem{hua2015microblog}
W.~Hua, K.~Zheng, and X.~Zhou, ``Microblog entity linking with social temporal
  context,'' in \emph{SIGMOD}, 2015, pp. 1761--1775.

\bibitem{zhang2020novel}
S.~Zhang, E.~Meij, K.~Balog, and R.~Reinanda, ``Novel entity discovery from web
  tables,'' in \emph{WWW}, 2020, pp. 1298--1308.

\bibitem{ritze2016profiling}
D.~Ritze, O.~Lehmberg, Y.~Oulabi, and C.~Bizer, ``Profiling the potential of
  web tables for augmenting cross-domain knowledge bases,'' in \emph{WWW},
  2016, pp. 251--261.

\bibitem{bhagavatula2015tabel}
C.~S. Bhagavatula, T.~Noraset, and D.~Downey, ``Tabel: Entity linking in web
  tables,'' in \emph{ISWC}, 2015, pp. 425--441.

\bibitem{tan2017entity}
C.~Tan, F.~Wei, P.~Ren, W.~Lv, and M.~Zhou, ``Entity linking for queries by
  searching wikipedia sentences,'' in \emph{EMNLP}, 2017, pp. 68--77.

\bibitem{blanco2015fast}
R.~Blanco, G.~Ottaviano, and E.~Meij, ``Fast and space-efficient entity linking
  for queries,'' in \emph{WSDM}, 2015, pp. 179--188.

\bibitem{cornolti2016piggyback}
M.~Cornolti, P.~Ferragina, M.~Ciaramita, S.~R{\"u}d, and H.~Sch{\"u}tze, ``A
  piggyback system for joint entity mention detection and linking in web
  queries,'' in \emph{WWW}, 2016, pp. 567--578.

\bibitem{wang2017named}
F.~Wang, W.~Wu, Z.~Li, and M.~Zhou, ``Named entity disambiguation for questions
  in community question answering,'' \emph{KBS}, vol. 126, pp. 68--77, 2017.

\bibitem{lin2020kbpearl}
X.~Lin, H.~Li, H.~Xin, Z.~Li, and L.~Chen, ``Kbpearl: a knowledge base
  population system supported by joint entity and relation linking,''
  \emph{Proc. VLDB Endow.}, vol.~13, no.~7, pp. 1035--1049, 2020.

\bibitem{liu2021joint}
Y.~Liu, W.~Shen, Y.~Wang, J.~Wang, Z.~Yang, and X.~Yuan, ``Joint open knowledge
  base canonicalization and linking,'' in \emph{SIGMOD}, 2021, pp. 2253--2261.

\bibitem{li2020survey}
J.~Li, A.~Sun, J.~Han, and C.~Li, ``A survey on deep learning for named entity
  recognition,'' \emph{TKDE}, 2020.

\bibitem{broscheit2019investigating}
S.~Broscheit, ``Investigating entity knowledge in bert with simple neural
  end-to-end entity linking,'' in \emph{CoNLL}, 2019, pp. 677--685.

\bibitem{liu2019roberta}
Y.~Liu, M.~Ott, N.~Goyal, J.~Du, M.~Joshi, D.~Chen, O.~Levy, M.~Lewis,
  L.~Zettlemoyer, and V.~Stoyanov, ``Roberta: A robustly optimized bert
  pretraining approach,'' \emph{arXiv preprint arXiv:1907.11692}, 2019.

\bibitem{joshi2020spanbert}
M.~Joshi, D.~Chen, Y.~Liu, D.~S. Weld, L.~Zettlemoyer, and O.~Levy, ``Spanbert:
  Improving pre-training by representing and predicting spans,'' \emph{TACL},
  vol.~8, pp. 64--77, 2020.

\bibitem{yang2019xlnet}
Z.~Yang, Z.~Dai, Y.~Yang, J.~Carbonell, R.~Salakhutdinov, and Q.~V. Le,
  ``Xlnet: generalized autoregressive pretraining for language understanding,''
  in \emph{NIPS}, 2019, pp. 5753--5763.

\bibitem{brown2020language}
T.~B. Brown, B.~Mann, N.~Ryder, M.~Subbiah, J.~Kaplan, P.~Dhariwal,
  A.~Neelakantan, P.~Shyam, G.~Sastry, A.~Askell \emph{et~al.}, ``Language
  models are few-shot learners,'' \emph{arXiv preprint arXiv:2005.14165}, 2020.

\end{thebibliography}

\vspace{-13mm}
\begin{IEEEbiography}[{\includegraphics[width=1in,height=1.25in,clip,keepaspectratio]{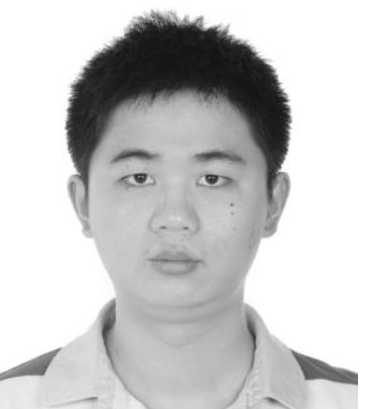}}]{Wei Shen} received the PhD degree in computer science from Tsinghua University, China, in 2014.
He is an associate professor in the College of Computer Science, Nankai University, China. His research interests include entity linking, knowledge base population, and text mining.
He was a recipient of ACM China Rising Star Award (Honorable Mention), CCF-Intel Young Faculty Researcher Program, and CAAI Outstanding Doctoral Dissertation Award.
\end{IEEEbiography}
\vspace{-13mm}

\begin{IEEEbiography}[{\includegraphics[width=1in,height=1.25in,clip,keepaspectratio]{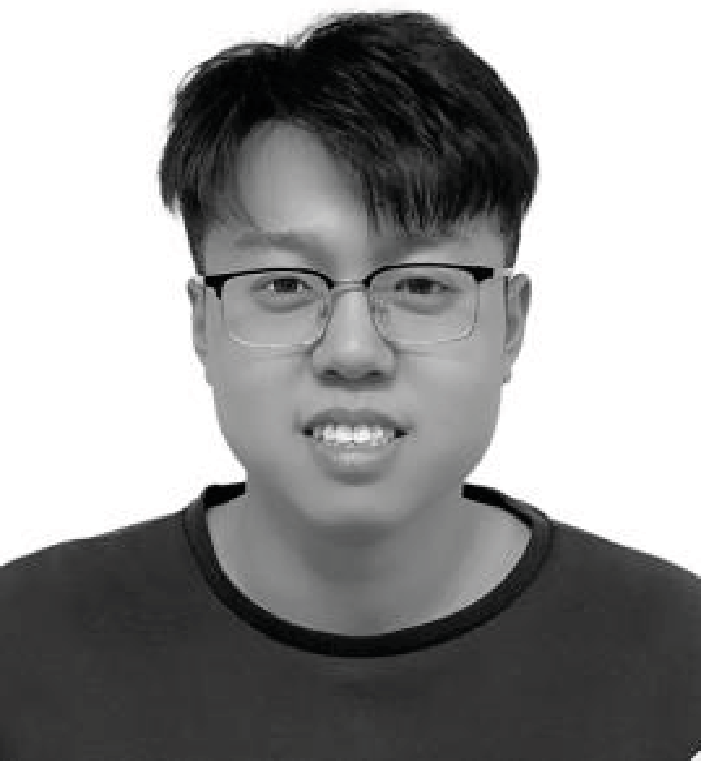}}]{Yuhan Li} received his BS degree from Northeast Forestry University, China in 2020. He is currently a master candidate at Nankai University. His research interests include knowledge graph, entity linking and data mining.

\end{IEEEbiography}
\vspace{-13mm}

\begin{IEEEbiography}[{\includegraphics[width=1in,height=1.25in,clip,keepaspectratio]{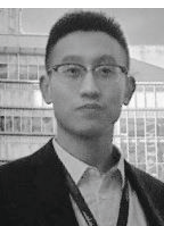}}]{Yinan Liu} received the BS degree from Northeastern University, China, in 2015 and the MS degree from Nankai University in 2018. He is a PhD student in the College of Computer Science in Nankai University. His research interests include knowledge base population and data mining.
\end{IEEEbiography}
\vspace{-13mm}

\begin{IEEEbiography}[{\includegraphics[width=1in,height=1.25in,clip,keepaspectratio]{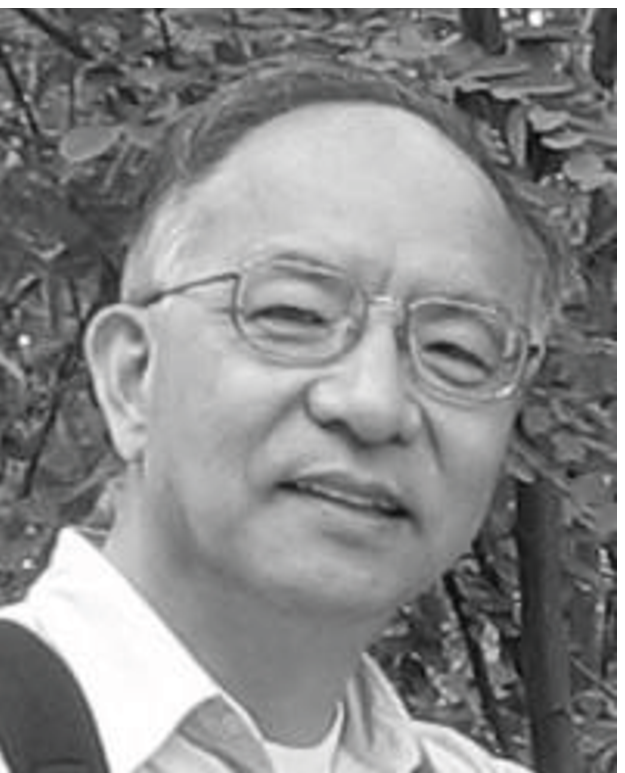}}]{Jiawei Han} is Michael Aiken Chair Professor in the Department of Computer Science, University of Illinois at Urbana-Champaign.  He received ACM SIGKDD Innovation Award (2004), IEEE Computer Society Technical Achievement Award (2005), IEEE Computer Society W. Wallace McDowell Award (2009), and Japan's Funai Achievement Award (2018). He is Fellow of ACM and Fellow of IEEE and served as the Director of Information Network Academic Research Center (INARC) (2009-2016) supported by the Network Science-Collaborative Technology Alliance (NS-CTA) program of U.S. Army Research Lab and co-Director of KnowEnG, a Center of Excellence in Big Data Computing (2014-2019), funded by NIH Big Data to Knowledge (BD2K) Initiative. 
\end{IEEEbiography}
\vspace{-13mm}

\begin{IEEEbiography}[{\includegraphics[width=1.25in,height=1.25in,clip,keepaspectratio]{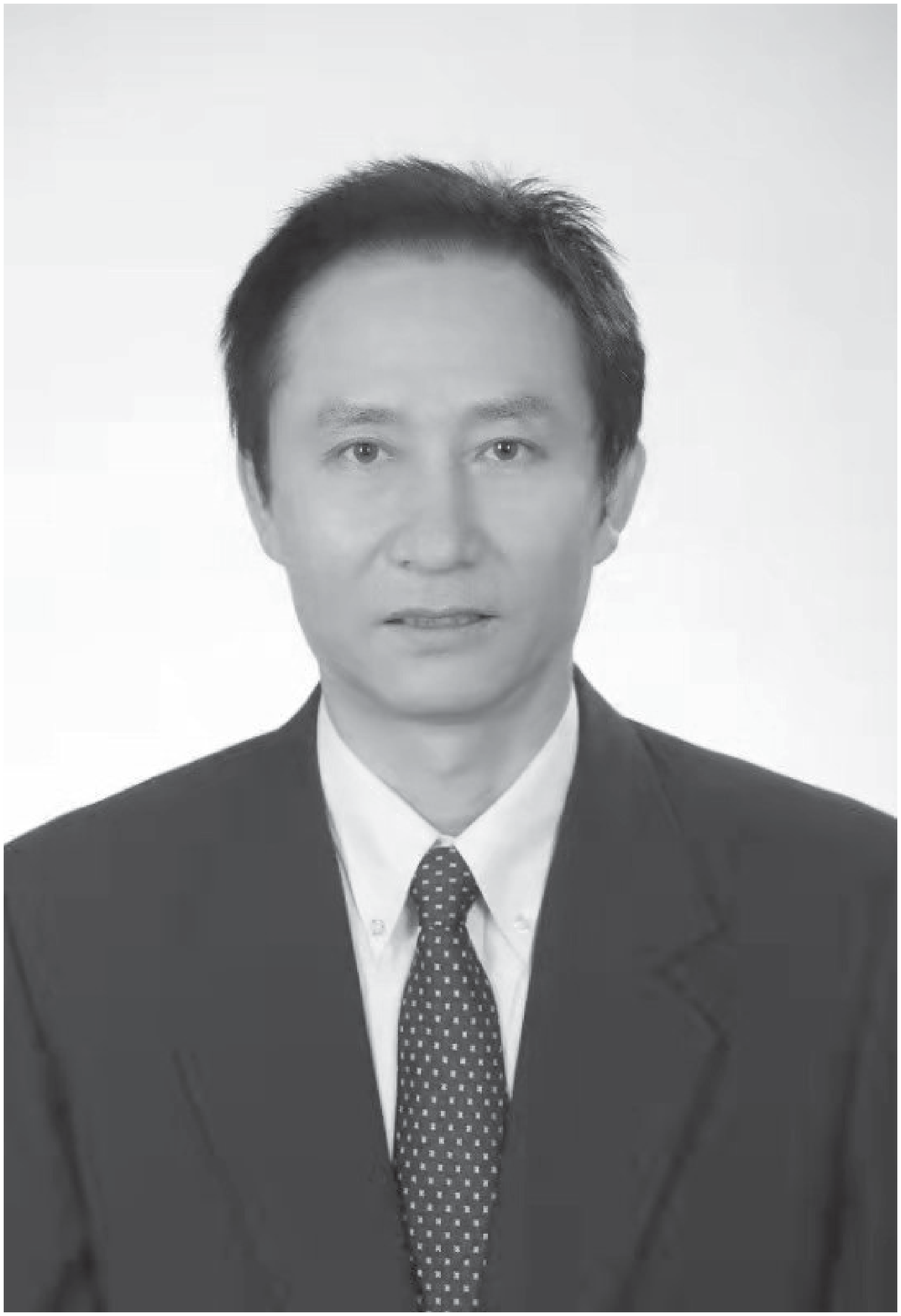}}]{Jianyong Wang}
 is currently a professor in the Department of Computer Science and Technology, Tsinghua University, Beijing, China.
 He received his PhD degree in Computer Science in 1999 from the Institute of Computing Technology, Chinese Academy of Sciences. His research interests mainly
 include data mining and Web information management. He is serving or ever served as a PC member for some leading international
 conferences, such as SIGKDD, VLDB, ICDE, WWW, and an associate editor of IEEE TKDE and ACM TKDD. He is a Fellow of the IEEE, a member
 of the ACM.
\end{IEEEbiography}

\vspace{-13mm}
\begin{IEEEbiography}[{\includegraphics[width=1in,height=1.25in,clip,keepaspectratio]{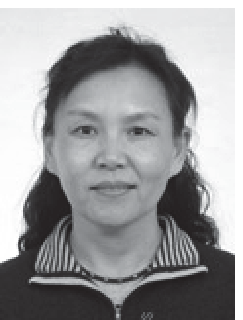}}]{Xiaojie Yuan} received the BS, MS, and the PhD
 degree in computer science from Nankai University.
 She is currently working as a professor of College of Computer Science, Nankai University.
 She leads a research group working on topics of database, data mining and information retrieval.
\end{IEEEbiography}

\end{document}